\documentclass[11pt]{article}

\usepackage[preprint]{acl}

\usepackage{times}
\usepackage{latexsym}

\usepackage[T1]{fontenc}

\usepackage[utf8]{inputenc}

\usepackage{microtype}

\usepackage{inconsolata}

\usepackage{graphicx}
\graphicspath{{./}}

\usepackage{enumitem}
\usepackage{amsmath,amssymb}
\usepackage{booktabs}
\usepackage{colortbl}
\usepackage{adjustbox}
\usepackage{bm}
\usepackage{siunitx}
\usepackage{array}
\usepackage{pifont}

\newcommand{\samethanks}{\footnotemark[1]}
\title{FederatedSkill: Federated Learning for Agentic Skill Evolution}

\author{
  \textbf{Jingbo Yang\textsuperscript{1}}\thanks{Equal contribution. Correspondence to: \texttt{jingbo@ucsb.edu, gyao@ucsb.com}}\quad
  \textbf{Guanyu Yao\textsuperscript{1}}\samethanks\quad
  \textbf{Yang Zhang\textsuperscript{2}}\quad
\\
  \textbf{Ramana Rao Kompella\textsuperscript{3}}\quad
  \textbf{Gaowen Liu\textsuperscript{3}}\quad
  \textbf{Shiyu Chang\textsuperscript{1}}
\\
\\
  \textsuperscript{1}UC Santa Barbara\quad
  \textsuperscript{2}MIT-IBM Watson AI Lab\quad
  \textsuperscript{3}Cisco Research
}

\usepackage{tcolorbox}
\tcbuselibrary{breakable,skins}
\begin{document}
\maketitle
\begin{abstract}
Modern LLM agents increasingly rely on skill libraries to handle complex tasks, making skill evolution a primary driver of self-improvement. However, isolated single-user task streams lack the diversity required to build comprehensive skills. While cross-user collaboration can overcome this data bottleneck, current trajectory-sharing approaches compromise user privacy and impose a uniform global library that fails to accommodate client heterogeneity. We introduce \textbf{FederatedSkill}, a privacy-preserving framework for collaborative agent evolution. Moving beyond raw trajectory sharing, FederatedSkill utilizes semantic \emph{skill diffs}, structured patches over local libraries, as the fundamental unit of communication. On the server side, an evolution agent aggregates these patches to dynamically model client-specific capability boundaries, facilitating strictly personalized skill evolution rather than a suboptimal global average. Evaluated across 20 distinct agent task families, \textbf{FederatedSkill} demonstrates substantial gains over self-evolving baselines, achieving up to a \textbf{44.4\%} increase in success rate and a \textbf{37.5\%} reduction in computational cost. Our code is available at \url{https://github.com/UCSB-NLP-Chang/FederatedSkill}.
\end{abstract}

\begin{figure*}[t]
    \centering
    \includegraphics[width=0.95\textwidth]{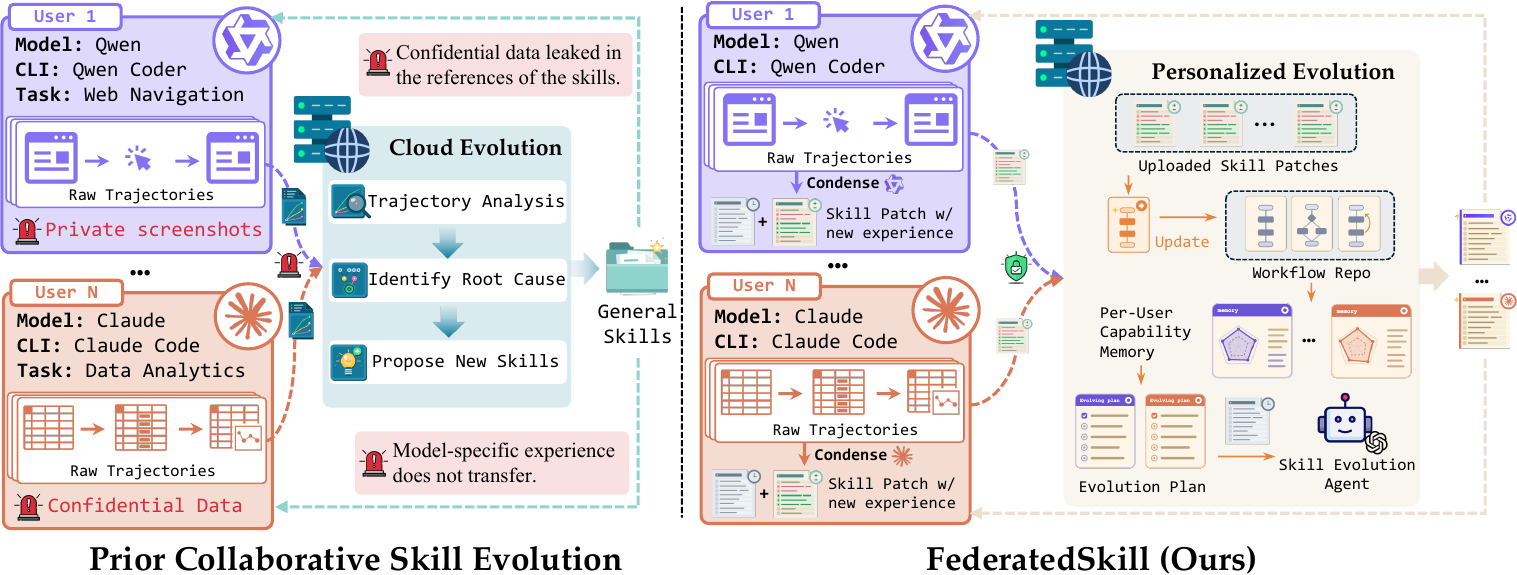}
    \caption{\textbf{Overview of FederatedSkill.} Each client agent self-reflects on its local trajectories and distills them into a structured \emph{skill patch} consisting of \textsc{add}, \textsc{edit}, or \textsc{delete} operations over the existing skill library, and uploads only the patch. A server-side \emph{merger agent} reads patches together with each client's agent profile and returns per-client personalized skill updates. Raw trajectories never leave the client.}
    \label{fig:teaser}
\end{figure*}

\section{Introduction}

Recently, the use of a library of \textit{skills} has become a mainstream paradigm for agentic systems, including the autonomous agent harnesses, such as OpenClaw and Claude Code. Agentic skills refer to reusable procedural memories that encode how to call tools, navigate environments, and chain multi-step actions~\citep{xu2026agent,ling2026agent,li2026skillsbench}, and can enable a great diversity of agentic tasks, ranging from short-form question answering to long-horizon web navigation, enterprise automation, and scientific tool use~\citep{liu2026well,merrill2026terminal,yang2026ares,yang2025webdart,patwardhan2025gdpval}.

In such paradigms, the primary way for agents to perform self-improvement is to \textit{evolve the skill library} \citep{ouyang2026skillos,zhang2026skillevolver,xia2026skillrl,shi2026skill1}. This involves adding new procedures, refining existing workflows, and pruning stale entries. Yet, a single user's task stream covers only a narrow slice of the skills they will eventually need. Consequently, agents often fail at new tasks before these experiences can be converted into skills, or they fail to learn from repeated attempts because they lack successful reference examples to provide guidance.

A natural way to break this single-user data bottleneck is to evolve skills \emph{collaboratively} across multiple users, which can expand the coverage of long-tail and challenging tasks. Recent collaborative skill-evolution frameworks~\citep{ma2026skillclaw} pursue this through a \emph{trajectory-sharing} paradigm: users upload their raw execution trajectories to a central server, which summarizes them into a shared skill library for redistribution. However, this paradigm faces two fundamental obstacles that no existing approach jointly resolves. \textbf{First}, raw trajectories often embed sensitive content (such as private information on the webpage, or confidential data in a spreadsheet) that users cannot safely transmit to an external server, even a trusted one. \textbf{Second}, a single global skill library is rarely optimal for everyone. Recent studies show that the most effective skill for a task depends on the user's specific model family, their agent framework, and their actual task distribution~\citep{li2026skillsbench,zhong2026skilllearnbench,liu2026well,wang2025promptbridge}. Consequently, naive aggregation yields a generalized library that underserves heterogeneous users. Together, these obstacles raise the central question: \emph{can users collaboratively evolve skills while preserving privacy, and still obtain a skill library that is tailored to their specific needs?}

In this paper, we propose \textbf{FederatedSkill}, a privacy-preserving, collaborative skill-evolution framework. To resolve the first challenge on privacy preservation, FederatedSkill avoids uploading the client's local trajectories, but instead lets the clients evolve their skills locally based on their own trajectories, and only uploads the resulting \emph{skill patch}, which documents add, edit, or delete operations applied to their skill libraries. This concept is analogous to federated learning, where the transmission of the sensitive training data is avoided by transmitting the local model weight updates~\citep{mcmahan2017communication}.
As illustrated in Figure~\ref{fig:teaser}, each client agent self-reflects on its own trajectories~\citep{shinn2023reflexion}, generates a skill patch, and uploads only that patch. A central server then merges these patches across clients (similar to merging concurrent git commits) and redistributes the updated library. Because the system exchanges semantically meaningful, user-reviewable patches rather than raw interaction logs, trajectories remain strictly local by construction. This design effectively eliminates the data-leakage vulnerabilities that the traditional trajectory-sharing paradigm cannot resolve.

To address the second challenge on client heterogeneity, FederatedSkill augments the server with an \emph{evolution agent} designed to achieve customized skill evolution. Rather than relying on a static profile, the evolution agent actively analyzes the skill patches uploaded by each client over time. Through these patch submissions, the evolution agent implicitly models the client's capability boundaries. Crucially, this modeling is continuously updated throughout the evolution process. Using this dynamic understanding, the evolution agent selects, adapts, and specializes skills from the global pool, projecting them into a personalized library tailored to that specific client. This approach places FederatedSkill alongside personalized federated learning~\citep{fallah2020personalized}, but introduces a crucial distinction. Unlike traditional federated learning that operates at the parameter level, FederatedSkill achieves personalization at the semantic-skill level via the evolution agent. This approach generalizes to a more heterogeneous client setup (\emph{i.e.}, different models and agent harnesses), which is unsupported by personalized federated learning.

We evaluate FederatedSkill on SkillFlow~\citep{zhang2026skillflow}, a recent benchmark for agent skill evolution spanning 20 distinct tasks. Compared to self-evolution, FederatedSkill consistently improves client agent performance, achieving up to a \textbf{44.4\%} gain in success rate and a \textbf{37.5\%} reduction in costs, all while strictly preserving client privacy.
Moreover, ablation studies highlight the necessity of personalization. By maintaining client-specific skill libraries rather than a single global library, FederatedSkill achieves an average performance gain of \textbf{12.2\%} across heterogeneous client settings.

\section{Related Work}
\label{sec:related-work}
\paragraph{Agent Self-Evolution.}
Agent self-evolution enables agents to autonomously improve through their own execution experiences. An early line of work focused on \emph{trajectory-level reflection}, where agents verbalize lessons from past attempts to serve as context for subsequent trials~\citep{shinn2023reflexion,zhang2024agent}. Building upon this, persistent experience stores convert past trajectories into retrievable insights~\citep{zhao2024expel}, structured memories~\citep{zhang2026memskill,zhang2026memrl,wang2024agent}, or transferable reasoning patterns~\citep{ouyang2025reasoningbank}. A more recent paradigm treats the \emph{skill library} itself as the locus of evolution, as semantic skills are inherently easier to reuse and generalize~\citep{ouyang2026skillos,zhang2026skillevolver,xia2026skillrl,shi2026skill1,yang2026autoskill}. However, these approaches predominantly follow a single-user paradigm, inevitably suffering from data scarcity. 
To bridge this gap, SkillClaw~\citep{ma2026skillclaw} explores multi-party skill evolution; yet, it suffers from privacy leakage by sharing raw trajectories and yields only a single, non-personalized global skill. FederatedSkill resolves both critical limitations by exchanging semantic skill patches instead of raw trajectories and executing per-client personalized evolution on the server. Furthermore, while Fed-SE~\citep{chen2025fed} also investigates federated agent evolution, its reliance on parameter-level LoRA aggregation severely limits its generalizability across heterogeneous clients.

\paragraph{Agent Skills.}
The skill paradigm organizes an agent's procedural knowledge into a library of reusable artifacts, retrieved and composed dynamically at runtime. While earlier research explores skills for lifelong agent learning~\citep{wang2023voyager,zheng2025skillweaver}, these approaches predominantly treat skills merely as in-context prompt components. More recently, agent harnesses such as Claude Code and OpenClaw have converged on filesystem-based skill specifications (\textit{e.g.}, \texttt{SKILL.md}). This shift elevates the skill library from a transient in-context scratchpad to a first-class deployment artifact. In parallel, emerging benchmarks measure how effectively agents \emph{utilize} a fixed skill set~\citep{li2026skillsbench,liu2026well} or \emph{evolve} one over a task sequence~\citep{zhang2026skillflow,zhong2026skilllearnbench}, firmly establishing the skill library as an independent unit of study.

\section{Problem Formulation}
\label{sec:problem-formulation}

We formalize cross-client skill evolution as a federated optimization problem over $N$ clients and $T$ rounds. At round $t$, each client $i$ holds a task distribution $\mathcal{D}_i$, a static profile $\rho_i$ (backbone model and agent harness), and a skill library $\mathcal{L}_i^t$. Raw trajectories and $\mathcal{D}_i$ never leave the client; only structured \emph{skill patches} are communicated.

\paragraph{Client-side Execution Agent.}
Client $i$ runs its agent policy $\pi_i(\cdot \mid \mathcal{L}_i^t, \rho_i)$ on tasks $x \sim \mathcal{D}_i$. The expected performance of a library $\mathcal{L}$ is:
\begin{equation}
    J_i(\mathcal{L}) = \mathbb{E}_{x \sim \mathcal{D}_i,\, \tau \sim \pi_i} \bigl[R_{i,x}(\tau)\bigr].
    \label{eq:objective_client}
\end{equation}
From its local trajectory batch $\mathcal{B}_i^t$, the client applies a reflection procedure $g_i$ to extract a semantic \emph{skill patch}:
\begin{equation}
    \delta_i^t = g_i\bigl(\mathcal{L}_i^t,\, \mathcal{B}_i^t,\, \rho_i\bigr),
    \label{eq:patch}
\end{equation}
where $\delta_i^t$ consists of high-level edits (e.g., \textsc{add}, \textsc{edit}, \textsc{delete}) to $\mathcal{L}_i^t$. By uploading only $\delta_i^t$ instead of $\mathcal{B}_i^t$, the framework preserves privacy by construction.

\paragraph{Server-side Evolution Agent.}
We cast the server-side aggregation as a Partially Observable Markov Decision Process (POMDP). At round $t$, the LLM-based evolution agent $\mathcal{M}$ receives only partial observations of the clients' true environments: $O^t = \{(\rho_i, \delta_i^t)\}_{i=1}^{N}$. Because raw trajectories are hidden, $\mathcal{M}$ must dynamically infer each client's underlying capability boundaries from the semantic content of $\delta_i^t$. Based on this inferred state, it generates a personalized update $\Delta_i^t$ for each client, yielding $\mathcal{L}_i^{t+1} = \mathrm{Apply}(\mathcal{L}_i^t, \Delta_i^t)$.

Given client weights $q_i$, the population performance is $\bar{J}^t = \sum_i q_i J_i(\mathcal{L}_i^t)$. The objective of $\mathcal{M}$ is to maximize the expected cumulative performance gain across $T$ rounds, subject to the constraint that it only observes $O^t$ at each round:
\begin{equation}
    \max_{\mathcal{M}}\ \mathbb{E} \left[ \sum_{t=0}^{T-1} \bigl(\bar{J}^{t+1} - \bar{J}^t\bigr) \right].
    \label{eq:objective_global}
\end{equation}

\section{FederatedSkill}
\label{sec:federatedskill}

FederatedSkill runs as a privacy-preserving client-server loop. Section~\ref{sec:method-client} covers the client side: local agents execute tasks and distill trajectories into skill patches. Section~\ref{sec:method-server} covers the server side: an evolution agent models each client's capabilities and emits personalized library updates.

\subsection{Client Execution and Patch Distillation}
\label{sec:method-client}

The client pipeline has two stages: local task execution collects a trajectory batch $\mathcal{B}_i^t$, then local patch distillation converts the batch into a skill patch $\delta_i^t$.

\subsubsection{Trial Execution}
\label{sec:method-trial}

During round $t$, client $i$ constructs the trajectory batch $\mathcal{B}_i^t$ by executing a set of sampled tasks $x \sim \mathcal{D}_i$. 
Specifically, the local agent is conditioned on its current skill library $\mathcal{L}_i^t$ and its specific LLM backbone. 
The agent autonomously searches the library and issues tool calls to generate an execution trajectory $\tau$. 
Each trajectory is then evaluated by the environment to assign a verification reward $R_{i,x}(\tau)$. 
Crucially, the base policy of the execution agent remains frozen across all federated rounds; thus, any performance gains stem strictly from the evolution of the skill library.

\subsubsection{Per-Client Patch Distillation}
\label{sec:method-patcher}

A local \emph{patcher}, which shares the same backbone LLM as the execution LLM, converts the raw trial artifacts into the skill patch $\delta_i^t$.

\paragraph{Patch Generation.}
The patcher generates $\delta_i^t$ via a single LLM call, based on three inputs:
\ding{182} \textit{Compacted Trajectory}: A compressed sequence retaining at most $K_{\text{step}}$ agentic steps (the initial step plus the $K_{\text{step}}-1$ most recent steps). Execution metrics are stripped, and environment observations are truncated to $K_{\text{obs}}$ characters via an explicit \texttt{<truncated>} marker.
\ding{183} \textit{Library Snapshot}: A JSON snapshot of the current library $\mathcal{L}_i^t$ detailing all file paths and their corresponding contents.
\ding{184} \textit{Trial Outcome}: Summary metadata including the task name, a quality signal $R_{i,x}(\tau)$, exception types, the final agent message, and verification sub-test failures (if available).

\paragraph{Patch Schema.}
The patcher outputs a structured four-tuple:
\begin{equation}
    \delta_i^t = \bigl(U_i^t, D_i^t, R_{i,x}(\tau), s_i^t\bigr),
\end{equation}
where each component is defined as follows:
\begin{itemize}[leftmargin=10pt,topsep=2pt,itemsep=1pt]
    \item $U_i^t$ is a dictionary of file \emph{upserts}, keyed by relative paths within the library and valued by full file contents. Each skill resides in a capability-named directory containing a \texttt{SKILL.md} file with frontmatter, alongside optional \texttt{scripts/}, \texttt{references/}, and \texttt{assets/} subfolders.
    \item $D_i^t$ is a list of relative paths proposed for deletion from $\mathcal{L}_i^t$.
    \item $s_i^t$ is a one-sentence natural language summary explaining the rationale behind the patch, recorded for downstream auditing.
    \item $R_{i,x}(\tau)$ is the task reward.
\end{itemize}

\paragraph{Privacy-Preserving Design.}
The patch $\delta_i^t$ is the only artifact that leaves the client. The raw trial artifacts stay on the local worker, including full trajectories, verifier output, intermediate files, and task inputs. Three design choices enforce this minimization. 
First, every field encodes only \emph{library-level} semantics: $U_i^t$ contains reusable playbooks intended to generalize across tasks (explicitly prompted to omit task-specific values, IDs, or one-off outputs), and $D_i^t$ contains only structural paths. No field carries raw interaction text. 
Second, because distillation runs entirely on the client's local backbone, the server only receives what the client model itself judges to be a generalizable lesson, never the raw trajectory that justified it. 
Third, as a system security measure, all paths in $U_i^t \cup D_i^t$ are rigorously validated to reject absolute paths or directory traversals. We provided an empirical privacy audit (Appendix~\ref{sec:privacy}) confirms these design choices.

\subsection{Server-Side Personalized Evolution}
\label{sec:method-server}

At the end of each round, the server collects the patch set $\{\delta_i^t\}_{i=1}^{N}$ from the $N$ clients of a task family and evolves each library individually. The evolution agent $\mathcal{M}$ acts under partial observability: it sees $O^t = \{(\rho_i, \delta_i^t)\}_{i=1}^{N}$ but never the raw trajectories, so it must infer each client's capability boundaries from the patch semantics alone. 

Personalization is necessary because a successful peer patch $\delta_j^t$ was validated against client $j$'s profile $\rho_j$, not $i$'s. Transplanting it wholesale re-introduces the cross-model/harness drift. The evolution objective is therefore not consensus but \emph{selective absorption}: a peer's success is evidence of a reusable workflow, not a template to copy. We instantiate $\mathcal{M}$ as a skill-driven agent that operates in two stages: Stage~1 transforms the round's patches into a per-client \emph{evolution plan} $\mathcal{P}^t$, and Stage~2 executes that plan to evolve the skill libraries.

\subsubsection{Stage 1: Evolution Planning}
\label{sec:method-stage1}

In Stage~1 the agent distills the round's patches into an evolution plan $\mathcal{P}^t$ with three components: a capability matrix locating each client within a specific kind of tasks, a two-level memory recording capability boundaries, and directives prescribing next steps.

\paragraph{Capability Matrix.}
The agent takes the full patch set together with a description-level digest of every client's pre-task library, restricted to skill names and descriptions, and produces a persistent \emph{capability matrix} $C^t$. Each row is a task \emph{workflow}, a distinct capability of the family characterized by its inputs, intermediate steps, and output schema. Each column is a client. The evolution agent cannot only add new rows if new task variants surface but also refines existing ones if they are not accurate.

Each cell in $C^t$ records how well a client has mastered a specific workflow, assigning one of four states: \emph{covered} (the client reliably solves it), \emph{absorbing} (this round's patch supplies the missing skill), \emph{broken} (a skill exists but fails, requiring repair), or \emph{gap} (no working skill exists). The agent determines this state by correlating the client's pre-task library with the skill patch, especially the trial reward $R_{i,x}(\tau)$. A workflow row is retired only when every client's cell becomes \emph{covered}, ensuring $C^t$ persistently tracks the capabilities the family has yet to collectively master.

\paragraph{Two-Level Capability Memory.}
While the matrix records \emph{whether} a client succeeds, the agent also maintains a persistent, two-level capability memory to record \emph{why} they fail. The \emph{high-level memory} is shared across the family: it summarizes global observations such as which workflows remain universally unsolved, orienting every client toward a common task frontier. The \emph{low-level memory} is private to each client and keyed by its profile $\rho_i$. It accumulates model-specific failure modes abstracted across rounds, such as a frequent misuse of a specific tool. 

The matrix and the two memory modules together form $\mathcal{M}$'s explicit model of each client's capability. They are the latent state that the POMDP in Equation~\ref{eq:objective_global} tracks. The server can then evolve each client around its own strengths and blind spots.

\paragraph{Evolution Directives.}
Directives translate the capability model into actionable instructions. A directive targets an open cell (a \emph{gap}, \emph{broken}, or \emph{absorbing} workflow) and prescribes how to close it: which skill to add, repair, or refactor, and the supporting evidence. These directives are grounded in both the matrix and the low-level memory, prescribing what to do and what to avoid. If a peer holds a verified skill but the low-level memory flags a style mismatch with client $i$'s backbone, the directive instructs the agent to refactor rather than copy the peer's logic. 

\subsubsection{Stage 2: Per-Client Library Evolution}
\label{sec:method-stage2}

\paragraph{Executing the Plan.}
Stage~2 runs independently per client to execute the directives in $\mathcal{P}^t$. For client $i$, the agent processes these directives alongside the current library $\mathcal{L}_i^t$, the round's incoming patches, and a read-only snapshot of peer libraries. It applies each directive to $\mathcal{L}_i^t$ by absorbing transferable patches, repairing broken skills in place, or rewriting peer skills to align with $\rho_i$.

\paragraph{Update and Carry-Forward.}
The agent directly modifies $\mathcal{L}_i^t$ to generate the personalized update $\Delta_i^t$, yielding the next-round library $\mathcal{L}_i^{t+1} = \mathrm{Apply}(\mathcal{L}_i^t, \Delta_i^t)$. This update is accompanied by an auditable decision log that records the source patch, reward, and justification for every modified path. Finally, the agent commits the current round's new observations to the low-level memory. This closes the evolution loop: the capability model continuously refines, ensuring that over $T$ rounds, the $N$ skill libraries advance toward a shared performance frontier while remaining strictly personalized to each client.

\section{Experiments}

\subsection{Experiment Setup}
\paragraph{Benchmark.}
We evaluate our method on the SkillFlow benchmark~\citep{zhang2026skillflow}, as this benchmark systematically evaluates the agent lifelong skill evolution. 
While there are other agent skill benchmark like SkillsBench~\citep{li2026skillsbench}, they mainly focus on evaluating agent's utilization of a \textbf{fixed} skill set, and only have one skill and one corresponding task for most tasks, which makes them not suitable for our setting.
SkillFlow consists of 20 diverse task families, each containing a series of sequential tasks arranged in increasing order of difficulty, which explicitly requires the evolution on the same skill.
Under this protocol, agents are initialized with an empty skill library and must tackle the tasks within each family sequentially. Following each round, the agent uses its past trajectories and reward signals to update its skill library. This process involves discovering, patching, transferring, or maintaining skills before the next round begins. Overall performance is evaluated based on the average task completion rate across the entire sequence.

\paragraph{Implementation.}
To reflect the real-world heterogeneity, we conduct our experiments using different backbone models, including Qwen3.6-Plus, GLM-5~\citep{zeng2026glm}, and Kimi K2.5~\citep{team2026kimi}, as well as different agent CLI, including Claude Code, Qwen Coder, and Kimi CLI. For all the inference hyper-parameters, we follow the default configuration. For our FederatedSkill framework, the server-side evolution agent is implemented using GLM-5 integrated with Claude Code, equipped with the skill described in Section~\ref{sec:method-server} to perform personalized evolution. The full evolution agent architecture is described in Appendix~\ref{app:artifacts}: a SKILL.md-driven \texttt{claude-code} agent with a two-step deterministic pipeline rather than a one-shot LLM call. On the client side, the self evolution is executed via local LLM calls to each client's designated backbone, following the schema in Section~\ref{sec:method-patcher}. We establish isolated \textit{agent self-evolution} as our non-collaborative baseline, wherein each client independently evolves its skill library relying solely on its own execution trajectories, without the influence of other clients.

\paragraph{Experiment Configuration.}
To assess collaborative skill evolution under increasing degrees of client
heterogeneity, we evaluate our method across three distinct setups.
(i) A \textit{homogeneous} setting, where all three clients share the same
backbone (GLM-5) and the same Harness (Claude Code).
(ii) A \textit{heterogeneous backbone} setting, where the three clients use different backbones (Qwen3.6-Plus, GLM-5, Kimi~K2.5) but still run on the identical Claude~Code harness.
(iii) A \textit{heterogeneous backbone/agent harness} setting, where each heterogeneous client
additionally operates within its own agent harness (Qwen Coder for Qwen,
Claude~Code for GLM, and Kimi-CLI for Kimi).

\begin{figure}[t]
  \centering
  \includegraphics[width=\columnwidth]{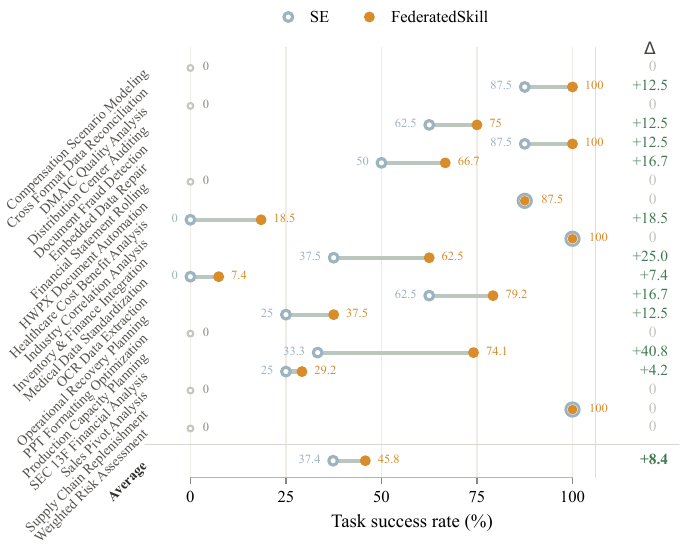}
  \caption{Per-task success rate of \textsc{FederatedSkill} compared to
  the self evolution baseline across 20 tasks with three \textbf{homogeneous GLM-5 clients}.}
  \label{fig:dumbbell}
\end{figure}

\begin{table*}[t]
\centering
\scriptsize
\setlength{\tabcolsep}{3pt}
\renewcommand{\arraystretch}{1.10}
\setlength{\arrayrulewidth}{0.5pt}

\definecolor{TopTint}{HTML}{E6ECF2}
\definecolor{KeyTint}{HTML}{F3F3F3}
\definecolor{PanelRule}{HTML}{B8B8B8}
\definecolor{ARESOrange}{HTML}{E07B00}
\definecolor{ARESTeal}{HTML}{1F8A8A}

\providecommand{\up}[1]{{\color{ARESOrange}$\uparrow$\,#1}}
\providecommand{\dn}[1]{{\color{ARESTeal}$\downarrow$\,#1}}
\providecommand{\nochg}{{\color{black!45}\textendash}}
\providecommand{\cli}[1]{\,\textsubscript{\texttt{#1}}}
\providecommand{\dsub}[1]{{\,\textsubscript{#1}}}
\providecommand{\avgskf}[2]{\multicolumn{1}{c}{\textbf{#1}\dsub{\textbf{#2}}}}

\sisetup{mode=text, detect-all=true}
\newcolumntype{E}{S[table-format=3.1, table-number-alignment=center]} 
\newcolumntype{K}{S[table-format=3.1, table-space-text-post={\dsub{\up{12.5}}}]} 

\resizebox{\textwidth}{!}{%
\begin{tabular}{l E K E K E K !{\color{PanelRule}\vrule width 0.6pt} E K E K E K}
\toprule
 & \multicolumn{6}{>{\columncolor{TopTint}[0pt][0pt]}c!{\color{PanelRule}\vrule width 0.6pt}}{\textsc{\textbf{Heterogeneous Backbone}}}
 & \multicolumn{6}{>{\columncolor{TopTint}[0pt][0pt]}c}{\textsc{\textbf{Heterogeneous Backbone \& Harness}}} \\
\cmidrule(lr){2-7}\cmidrule(lr){8-13}
 & \multicolumn{2}{c}{\textbf{Qwen3.6-Plus}\cli{Claude Code}}
 & \multicolumn{2}{c}{\textbf{GLM-5}\cli{Claude Code}}
 & \multicolumn{2}{c!{\color{PanelRule}\vrule width 0.6pt}}{\textbf{Kimi K2.5}\cli{Claude Code}}
 & \multicolumn{2}{c}{\textbf{Qwen3.6-Plus}\cli{Qwen Coder}}
 & \multicolumn{2}{c}{\textbf{GLM-5}\cli{Claude Code}}
 & \multicolumn{2}{c}{\textbf{Kimi K2.5}\cli{Kimi CLI}} \\
\cmidrule(lr){2-3}\cmidrule(lr){4-5}\cmidrule(lr){6-7}\cmidrule(lr){8-9}\cmidrule(lr){10-11}\cmidrule(lr){12-13}
\textbf{Task Family}
 & {SE} & {FedSkill} & {SE} & {FedSkill} & {SE} & {FedSkill}
 & {SE} & {FedSkill} & {SE} & {FedSkill} & {SE} & {FedSkill} \\
\midrule
Cross-Format-Data-Reconciliation & 100.0 & 100.0\dsub{\nochg}    & 87.5  & 87.5\dsub{\nochg}     & 87.5  & 100.0\dsub{\up{12.5}} & 100.0 & 100.0\dsub{\nochg}    & 87.5  & 100.0\dsub{\up{12.5}} & 87.5  & 75.0\dsub{\dn{12.5}}  \\
Distribution-Center-Auditing  & 62.5  & 75.0\dsub{\up{12.5}}  & 62.5  & 62.5\dsub{\nochg}     & 75.0  & 50.0\dsub{\dn{25.0}}  & 62.5  & 75.0\dsub{\up{12.5}}  & 62.5  & 50.0\dsub{\dn{12.5}}  & 37.5  & 62.5\dsub{\up{25.0}}  \\
Document-Fraud-Detection  & 87.5  & 100.0\dsub{\up{12.5}} & 87.5  & 87.5\dsub{\nochg}     & 75.0  & 75.0\dsub{\nochg}     & 100.0 & 100.0\dsub{\nochg}    & 87.5  & 100.0\dsub{\up{12.5}} & 87.5  & 100.0\dsub{\up{12.5}} \\
Embedded-Data-Repair  & 50.0  & 50.0\dsub{\nochg}     & 50.0  & 50.0\dsub{\nochg}     & 37.5  & 62.5\dsub{\up{25.0}}  & 50.0  & 50.0\dsub{\nochg}     & 50.0  & 50.0\dsub{\nochg}     & 37.5  & 37.5\dsub{\nochg}     \\
HWPX-Document-Automation  & 62.5  & 87.5\dsub{\up{25.0}}  & 87.5  & 75.0\dsub{\dn{12.5}}  & 62.5  & 87.5\dsub{\up{25.0}}  & 75.0  & 100.0\dsub{\up{25.0}} & 87.5  & 87.5\dsub{\nochg}     & 87.5  & 87.5\dsub{\nochg}     \\
Healthcare-Cost-Benefit-Analysis & 22.2  & 33.3\dsub{\up{11.1}}  & 0.0   & 22.2\dsub{\up{22.2}}  & 11.1  & 22.2\dsub{\up{11.1}}  & 66.7  & 55.6\dsub{\dn{11.1}}  & 0.0   & 11.1\dsub{\up{11.1}}  & 33.3  & 66.7\dsub{\up{33.3}}  \\
Industry-Correlation-Analysis  & 62.5  & 87.5\dsub{\up{25.0}}  & 100.0 & 100.0\dsub{\nochg}    & 87.5  & 100\dsub{\up{12.5}}     & 100.0 & 100.0\dsub{\nochg}    & 100.0 & 100.0\dsub{\nochg}    & 100.0 & 100.0\dsub{\nochg}    \\
Inventory-\&-Finance-Integration  & 50.0  & 50.0\dsub{\nochg}     & 37.5  & 62.5\dsub{\up{25.0}}  & 62.5  & 50.0\dsub{\dn{12.5}}  & 37.5  & 62.5\dsub{\up{25.0}}  & 37.5  & 50.0\dsub{\up{12.5}}  & 62.5  & 50.0\dsub{\dn{12.5}}  \\
Medical-Data-Standardization  & 0.0   & 11.1\dsub{\up{11.1}}  & 0.0   & 0.0\dsub{\nochg}      & 22.2  & 11.1\dsub{\dn{11.1}}  & 22.2  & 22.2\dsub{\nochg}     & 0.0   & 11.1\dsub{\up{11.1}}  & 0.0   & 11.1\dsub{\up{11.1}}  \\
OCR-Data-Extraction  & 50.0  & 50.0\dsub{\nochg}     & 62.5  & 75.0\dsub{\up{12.5}}  & 62.5  & 75.0\dsub{\up{12.5}}  & 75.0  & 62.5\dsub{\dn{12.5}}  & 62.5  & 75.0\dsub{\up{12.5}}  & 62.5  & 62.5\dsub{\nochg}     \\
Operational-Recovery-Planning  & 25.0  & 37.5\dsub{\up{12.5}}  & 25.0  & 37.5\dsub{\up{12.5}}  & 25.0  & 37.5\dsub{\up{12.5}}     & 50.0  & 37.5\dsub{\dn{12.5}}   & 25.0  & 37.5\dsub{\up{12.5}}  & 37.5  & 25.0\dsub{\dn{12.5}}  \\
Production-Capacity-Planning  & 66.7  & 77.8\dsub{\up{11.1}}  & 33.3  & 66.7\dsub{\up{33.3}}  & 11.1  & 44.4\dsub{\up{33.3}}  & 77.8  & 66.7\dsub{\dn{11.1}}  & 33.3  & 66.7\dsub{\up{33.3}}  & 44.4  & 88.9\dsub{\up{44.4}}  \\
SEC-13F-Financial-Analysis & 37.5  & 37.5\dsub{\nochg}     & 25.0  & 50.0\dsub{\up{25.0}}  & 12.5  & 25.0\dsub{\up{12.5}}  & 25.0  & 50.0\dsub{\up{25.0}}  & 25.0  & 37.5\dsub{\up{12.5}}  & 12.5  & 12.5\dsub{\nochg}     \\
Supply-Chain-Replenishment  & 66.7  & 77.8\dsub{\up{11.1}}  & 100.0 & 100.0\dsub{\nochg}    & 100.0 & 88.9\dsub{\dn{11.1}}  & 77.8  & 88.9\dsub{\up{11.1}}  & 100.0 & 100.0\dsub{\nochg}    & 88.9  & 100.0\dsub{\up{11.1}}  \\
PPT-Formatting-Optimization  & 0.0   & 0.0\dsub{\nochg}      & 0.0   & 0.0\dsub{\nochg}      & 0.0   & 0.0\dsub{\nochg}      & 0.0   & 12.5\dsub{\up{12.5}}  & 0.0   & 0.0\dsub{\nochg}      & 0.0   & 0.0\dsub{\nochg}      \\
\midrule
\rowcolor{KeyTint}[\tabcolsep][\tabcolsep]
\textbf{Avg. (\%)}
 & \multicolumn{1}{c}{36.75} & \avgskf{43.37}{\up{6.63}}
 & \multicolumn{1}{c}{37.35} & \avgskf{43.37}{\up{6.02}}
 & \multicolumn{1}{c}{36.14} & \multicolumn{1}{c!{\color{PanelRule}\vrule width 0.6pt}}{\textbf{40.96}\dsub{\textbf{\up{4.82}}}}
 & \multicolumn{1}{c}{45.78} & \avgskf{48.80}{\up{3.01}}
 & \multicolumn{1}{c}{37.35} & \avgskf{43.37}{\up{6.02}}
 & \multicolumn{1}{c}{38.55} & \avgskf{43.98}{\up{5.42}} \\
\bottomrule
\end{tabular}}
\vspace{-5pt}
\caption{\textbf{Main results on \textsc{FederatedSkill}.} 
\emph{Left panel (Heterogeneous Backbone):} All backbones operate on a uniform Claude Code harness. 
\emph{Right panel (\dots \& Harness):} Each backbone pairs with its native harness (Qwen Coder, Claude Code, Kimi-CLI). 
Subscripts denote the absolute performance change ($\Delta = \mathrm{FedSkill} - \mathrm{SE}$) in percentage points, highlighted by \textcolor{ARESOrange}{$\uparrow$} for gains and \textcolor{ARESTeal}{$\downarrow$} for regressions. 
Five consistently failing task families (scoring $0.0$ across all settings) are omitted for brevity: Compensation-Scenario-Modeling, DMAIC-Quality-Analysis, Financial-Statement-Rolling, Sales-Pivot-Analysis, and Weighted-Risk-Assessment.}
\label{tab:mixed-cli-se-vs-skf}
\end{table*}

\subsection{Main Results}

\paragraph{Homogeneous Clients.}
In the homogeneous setting, all three clients and the server-side evolution agent share an identical configuration: GLM-5 as the backbone and the \texttt{Claude Code} CLI as the agent harness.
As illustrated in Figure~\ref{fig:dumbbell}, FederatedSkill increases the average task success rate from 37.4\% to 45.8\%, confirming that the patch-sharing protocol effectively introduces complementary skills even without client heterogeneity. The most substantial improvements occur in Production Capacity Planning ($\uparrow 40.8$ pp), Inventory \& Finance Integration ($\uparrow 25.0$ pp), and Healthcare Cost Benefit Analysis ($\uparrow 18.5$ pp). In these domains, aggregating diverse experiences across clients allows the evolution agent to resolve specific blind spots missed by individual clients. In addition, both settings achieve 0\% accuracy on six tasks and 100\% on two others, indicating that performance on these specific tasks is bottlenecked by the backbone model's inherent capability ceiling rather than the evolution method.

\paragraph{Heterogeneous Backbone.}
As Table~\ref{tab:mixed-cli-se-vs-skf} demonstrates, \textsc{FederatedSkill} consistently improves performance across heterogeneous backbones under a uniform Claude Code harness. Average success rates increase for Qwen3.6-Plus ($36.75\% \to 43.37\%$), GLM-5 ($37.35\% \to 43.37\%$), and Kimi~K2.5 ($36.14\% \to 40.96\%$), achieving a robust average gain of $+5.82$\,pp. Cell-level analysis reveals two critical insights. 

First, \textsc{FederatedSkill} facilitates \emph{universal skill transfer} across distinct architectures. In Production-Capacity-Planning, all three models improve simultaneously ($+11.1$, $+33.3$, and $+33.3$\,pp, respectively). Similarly, in Healthcare-Cost-Benefit-Analysis, GLM-5 achieves a ``zero-to-one'' breakthrough ($0.0\% \to 22.2\%$), proving that the evolution server can successfully salvage a failing client by injecting semantic patches from peer models. 
Second, we observe localized regressions in a few specialized workflows, such as Kimi~K2.5 on Distribution-Center-Auditing ($75.0\% \to 50.0\%$) and Medical-Data-Standardization ($22.2\% \to 11.1\%$). This implies that cross-model patch evolution can occasionally induce negative transfer, highlighting a natural trade-off between broad heterogeneous generalization and model-specific reliability.

\paragraph{Heterogeneous Backbone \& Harness.}
In this extreme setting, clients differ not only in their backbone models but also in their native agent harnesses. Consequently, the server-side evolution agent must assimilate skill patches authored under different execution interfaces and tool-use conventions. Despite this additional heterogeneity, the right panel of Table~\ref{tab:mixed-cli-se-vs-skf} shows that FederatedSkill improves the overall average performance of every client. Specifically, Qwen3.6-Plus improves from 45.78\% to 48.80\%, GLM-5 from 37.35\% to 43.37\%, and Kimi~K2.5 from 38.55\% to 43.98\%, yielding an average gain of $+4.82$\,pp across the three clients. This gain is slightly below the single-heterogeneity setting ($+5.82$\,pp), indicating that cross-harness incompatibility introduces additional friction but does not eliminate the benefit of federation. FederatedSkill therefore remains effective even when clients operate under distinct harnesses, suggesting that the capability-level abstractions captured by the semantic patches ($\delta_i^t$) can generalize beyond the specific environments that generated them. Appendix~\ref{app:case-M} dissects the largest cell-level gain in this setting, where Kimi~K2.5 improves on Production-Capacity-Planning by $+44.4$\,pp. Appendix~\ref{app:case-regressions} analyzes representative regressions to illustrate edge-case failure modes of cross-harness patch merging.

\begin{figure}[t]
  \centering
  \includegraphics[width=0.95\columnwidth]{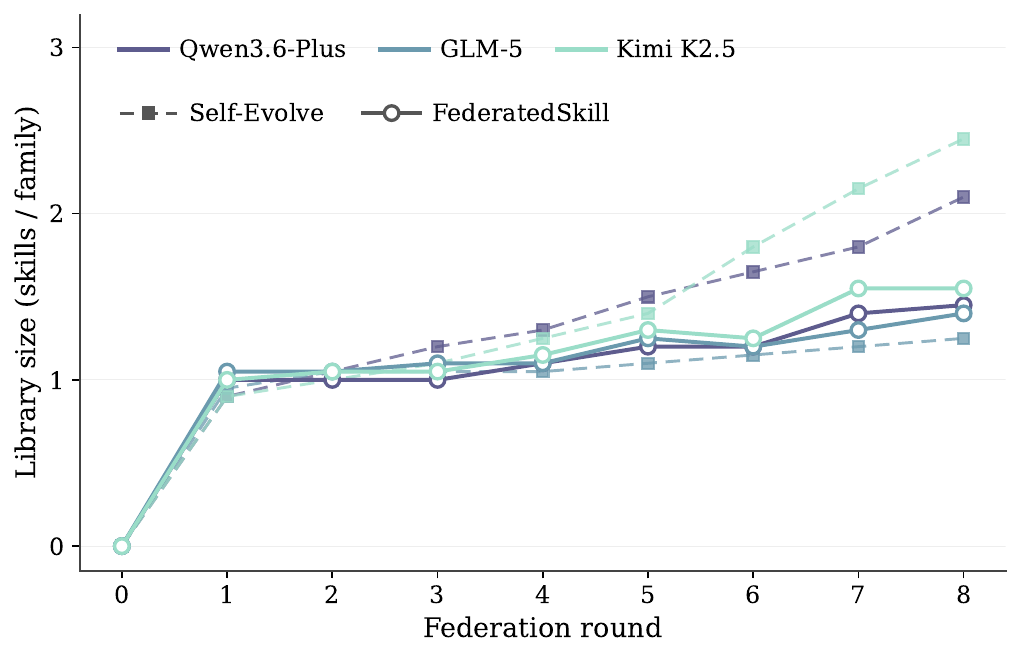}
  \caption{\textbf{Skill library size across federation rounds.}
  Number of distinct skills per family that each client carries into the next round. 
  }
  \label{fig:library-size}
\end{figure}

\subsection{Evolution Dynamics}
To understand the source of these performance improvements, we examine the evolution of the client-side skill library across federation rounds. As Figure~\ref{fig:library-size} illustrates, Self-Evolve (SE) exhibits diverging trends depending on the underlying model. For Kimi~K2.5 and Qwen3.6-Plus, SE suffers from \emph{library bloat}: clients continually generate new skills without adequate consolidation, yielding approximately 2.5 and 2.1 skills per task family by round~8, respectively. This unchecked proliferation increases the risk of retrieving semantically overlapping skills at runtime. Conversely, GLM-5 under SE exhibits \emph{library stagnation}, maintaining a nearly flat trajectory of 1.0 to 1.3 skills per family, thereby limiting its capacity to specialize across diverse task variants.

FederatedSkill successfully mitigates both extremes, constraining the libraries of all heterogeneous backbones within a tight band of roughly 1.0 to 1.6 skills per family throughout all 8 rounds. This bounded growth is driven by two key mechanisms within the server-side evolution agent: (i)~\emph{cross-client deduplication}, which merges semantically similar skills from disparate clients into a unified abstraction, and (ii)~\emph{conservative admission}, which integrates a new skill only when it represents a genuinely novel capability. By maintaining a highly consolidated library, FederatedSkill ensures that skill retrieval remains accurate and manageable even as heterogeneity scales, directly contributing to the performance gains reported in the \emph{Heterogeneous Backbone} panel of Table~\ref{tab:mixed-cli-se-vs-skf}. Appendix~\ref{app:case-B} details the per-round library compositions underlying these aggregate trajectories.

\begin{figure}[t]
  \centering
  \includegraphics[width=\columnwidth]{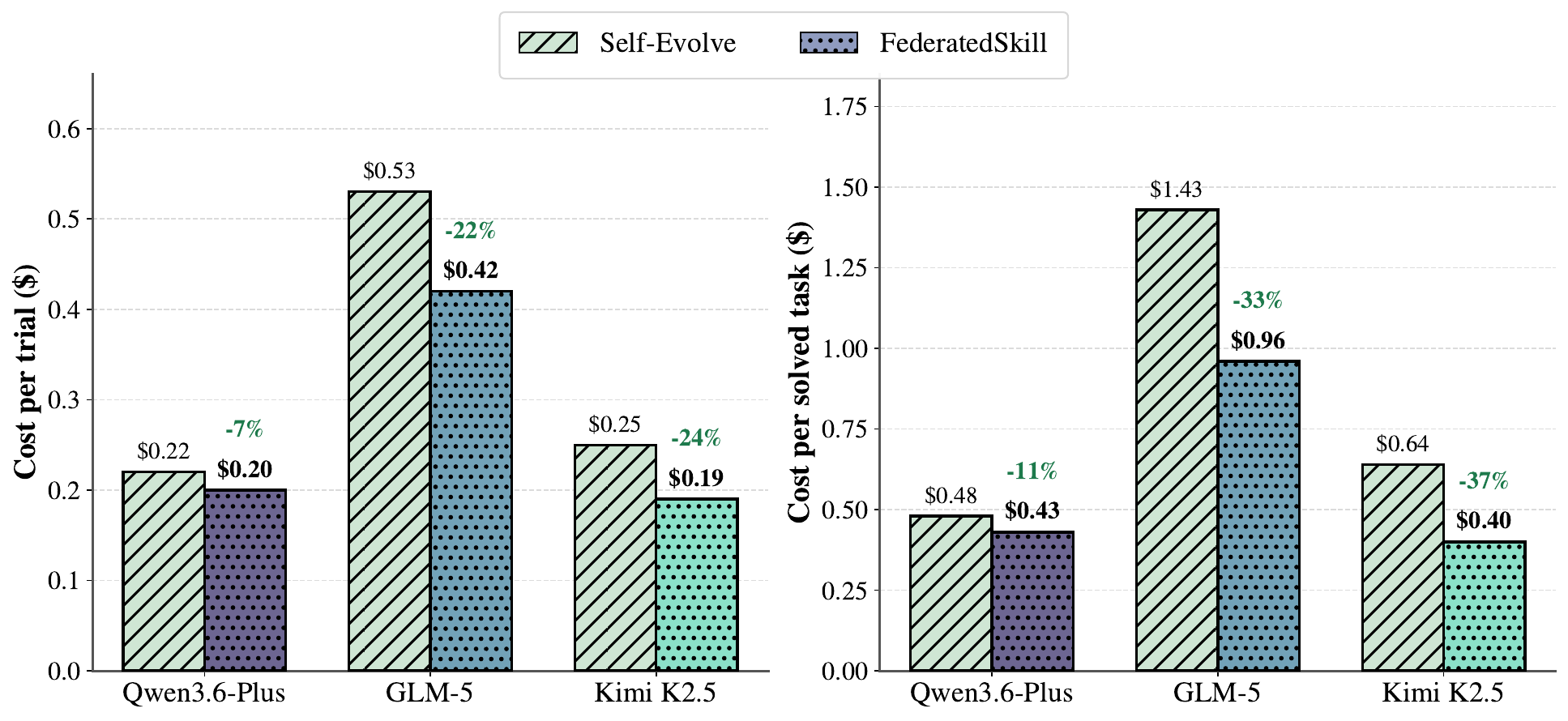}
  
  \caption{\textbf{Client API cost in the \textit{Heterogeneous Backbone \& Harness} setting.}
  Costs are computed using native non-cached list prices in USD.
  }
  \label{fig:cost}
\end{figure}

\subsection{Cost Analysis}
Figure~\ref{fig:cost} illustrates the API costs under the \emph{Heterogeneous Backbone \& Harness} setting, evaluated across the full 20-family pool. Costs are calculated directly from logged input and output token usage based on each provider's native list prices. To provide a comprehensive evaluation, we report both the average per-trial cost and the normalized cost per solved task.

As shown, \textsc{FederatedSkill} reduces the per-trial cost across all three backbones, by $7\%$ for Qwen3.6-Plus, $22\%$ for GLM-5, and $24\%$ for Kimi~K2.5. This reduction is primarily achieved by avoiding the excessive prompt growth associated with SE's library bloat, with the largest savings appearing on the two heavier backbones. Crucially, when normalizing costs by successful outcomes, the efficiency gap widens: \textsc{FederatedSkill} lowers the per-solved-task cost by $11\%$ for Qwen, $33\%$ for GLM-5, and $37\%$ for Kimi. Coupled with the performance gains in Table~\ref{tab:mixed-cli-se-vs-skf}, these findings demonstrate that \textsc{FederatedSkill} delivers substantially better cost-efficiency across diverse backbones without compromising task success. Beyond inference, we also prove the federated protocol itself is bandwidth-light in Appendix~\ref{app:communication_cost}.

\subsection{Ablation Study}
\label{sec:ablation}
We isolate the efficacy of the personalized evolution agent by comparing it against a non-personalized baseline (Global). While this baseline retains the patch-sharing protocol, it generates a single, uniform skill per task family for all clients, completely omitting per-client memory and personalized absorption. We evaluate on five task families drawn from diverse domains in the benchmark: Production-Capacity-Planning, Operational-Recovery-Planning, SEC-13F-Financial-Analysis, Industry-Correlation-Analysis, and OCR-Data-Extraction.

As detailed in Table~\ref{tab:ablation}, the personalized evolution agent improves aggregate performance for every client by $+9.8$ to $+12.2$\,pp. Kimi~K2.5 achieves the highest lift ($+12.2$\,pp), while Qwen3.6-Plus and GLM-5 both improve by $+9.8$\,pp. These results empirically validate that maintaining a single global skill inherently fails to accommodate the distinct reasoning boundaries of heterogeneous backbones, underscoring the necessity of per-client personalization.

\begin{table}[t]
\centering
\scriptsize
\setlength{\tabcolsep}{4.5pt}
\renewcommand{\arraystretch}{1.10}
\setlength{\arrayrulewidth}{0.5pt}
\definecolor{KeyTint}{HTML}{F3F3F3}
\definecolor{ARESOrange}{HTML}{E07B00}
\definecolor{ARESTeal}{HTML}{1F8A8A}
\providecommand{\up}[1]{{\color{ARESOrange}$\uparrow$\,#1}}
\providecommand{\dn}[1]{{\color{ARESTeal}$\downarrow$\,#1}}
\providecommand{\nochg}{{\color{black!45}\textendash}}
\resizebox{\columnwidth}{!}{%
\begin{tabular}{l cc cc cc}
\toprule
 & \multicolumn{2}{c}{\textbf{Qwen3.6-Plus}}
 & \multicolumn{2}{c}{\textbf{GLM-5}}
 & \multicolumn{2}{c}{\textbf{Kimi K2.5}} \\
\cmidrule(lr){2-3}\cmidrule(lr){4-5}\cmidrule(lr){6-7}
\textbf{Task Family}    & Global   & FedSkill            & Global   & FedSkill            & Global   & FedSkill            \\
\midrule
Production Planning &  55.6 &  77.8 &  55.6 &  66.7 &  33.3 &  44.4  \\
Recovery Planning &  12.5 &  37.5 &  37.5 &  37.5 &  12.5 & 37.5  \\
Financial Analysis &  25.0 &  37.5 &  25.0 &  50.0 &  25.0 &  25.0  \\
Industry Correlation &  75.0 &  87.5 & 100.0 & 100.0 &  75.0 &  100.0  \\
OCR Extraction &  75.0 &  50.0 &  62.5 &  75.0 &  75.0 &  75.0  \\
\midrule
\rowcolor{KeyTint}[\tabcolsep][\tabcolsep]
\textbf{Avg. (\%)} &  48.8 & \textbf{58.5}\,\textbf{\up{9.8}} &  56.1 & \textbf{65.9}\,\textbf{\up{9.8}} &  43.9 & \textbf{56.1}\,\textbf{\up{12.2}} \\
\bottomrule
\end{tabular}}
\caption{\textbf{Ablation: personalized merging vs.\ one global skill.}
We compare \textsc{FederatedSkill} (FedSkill, personalized evolution agent with per-client
memory $\mathcal{M}_i^t$) against a stripped-down baseline (Global) where the
evolution agent writes a single global skill per family.}
\label{tab:ablation}
\end{table}

\section{Conclusion}
\label{sec:conclusion}
We presented \textbf{FederatedSkill}, a privacy-preserving framework for collaborative agent evolution. By exchanging semantic \emph{skill patches} rather than raw trajectories, our approach safeguards client privacy while a server-side evolution agent curates personalized skill libraries tailored to individual capability boundaries. Evaluated on the SkillFlow benchmark, FederatedSkill improves task success by up to $44.4\%$ and reduces costs by $37.5\%$ over self-evolution baselines, demonstrating robust gains across highly heterogeneous clients. These results establish the skill library as a highly effective unit of federation for agent self-improvement.

\section*{Limitations}
While FederatedSkill demonstrates the efficacy of collaborative agent evolution at the semantic-skill level, it has a few limitations that present exciting avenues for future research. First, as this work primarily focuses on establishing the federated skill paradigm, we leave system-level optimizations---such as developing more bandwidth-efficient schemas for patch transmission---to future work. Additionally, our current framework assumes a trusted federated environment. Defending against adversarial clients attempting to upload malicious or poisoned skill patches goes beyond the scope of this paper, but is a critical next step for deploying such systems in the wild. 

Second, due to the substantial computational resources required to simulate full LLM-driven agent evolution loops, our empirical evaluation is constrained to a modest number of concurrent clients. While our results clearly validate the advantages of personalized patch aggregation, exploring the scalability and emergent behaviors of FederatedSkill at a massive scale (e.g., coordinating tens or hundreds of heterogeneous agents) remains an important direction for future research given expanded computational budgets.

\section*{Acknowledgement}
The work of Jingbo Yang, Guanyu Yao and Shiyu Chang was partially supported by National Science Foundation (NSF) Grant IIS-2338252, and NSF Grant IIS-2302730.

\appendix

\sloppy
\makeatletter
\renewcommand{\verbatim@font}{\scriptsize\ttfamily}
\makeatother


\newtcolorbox{promptbox}[1][]{
  colback=blue!3!white, colframe=blue!45!black,
  coltitle=white, colbacktitle=blue!45!black,
  boxrule=0.5pt, arc=2pt, left=4pt, right=4pt, top=2pt, bottom=2pt,
  before skip=6pt, after skip=6pt, breakable,
  fonttitle=\sffamily\bfseries\small, #1
}

\newtcolorbox{memorybox}[1][]{
  colback=teal!3!white, colframe=teal!50!black,
  coltitle=white, colbacktitle=teal!50!black,
  boxrule=0.5pt, arc=2pt, left=4pt, right=4pt, top=2pt, bottom=2pt,
  before skip=6pt, after skip=6pt, breakable,
  fonttitle=\sffamily\bfseries\small, #1
}

\newtcolorbox{databox}[1][]{
  colback=purple!3!white, colframe=purple!45!black,
  coltitle=white, colbacktitle=purple!45!black,
  boxrule=0.5pt, arc=2pt, left=4pt, right=4pt, top=2pt, bottom=2pt,
  before skip=6pt, after skip=6pt, breakable,
  fonttitle=\sffamily\bfseries\small, #1
}

\newtcolorbox{logbox}[1][]{
  colback=orange!4!white, colframe=orange!50!black,
  coltitle=white, colbacktitle=orange!50!black,
  boxrule=0.5pt, arc=2pt, left=4pt, right=4pt, top=2pt, bottom=2pt,
  before skip=6pt, after skip=6pt, breakable,
  fonttitle=\sffamily\bfseries\small, #1
}

\section{Per-Cell Case Analyses}\label{app:cases}

This section investigates the trajectory- and patch-level mechanisms driving the performance differences in Table~\ref{tab:mixed-cli-se-vs-skf}, focusing strictly on the \emph{Heterogeneous Backbone \& Harness} setting.

\subsection{Trajectory-Level Comparison on a Shared Task}\label{app:case-M}

To understand why Self-Evolution (SE) and \textsc{FederatedSkill} yield divergent outcomes under identical conditions, we isolate a matched pair: two Kimi~K2.5 agents tackling the exact same Production-Capacity-Planning (PCP) task in Round~6. The SE agent relies on its independently evolved library, while the \textsc{FederatedSkill} agent uses the library merged from the three-client federation.

\paragraph{Task Definition and SE Failure Mode.}
The task requires the agent to generate a 49-week capacity plan. The critical constraint is the initial condition at Week 3, defined as: $\text{Start of Week Past Due} + \text{Scheduled Demand} = 598.24$. The agent with SE skills fails because it assigns the combined total ($598.24$) entirely to $\text{Start of Week Past Due}$ while simultaneously assigning an additional $98.24$ to $\text{Scheduled Demand}$. This double-counts the demand, immediately triggering a verifier rejection.

Inspection of the SE agent's \texttt{SKILL.md} reveals the root cause of this failure. Rather than providing a formula for how to split the combined initial value, the skill merely defines a passive, post-generation checklist item (i.e., ``verify that start backlog $+$ demand $=$ expected total''). Without explicit calculation instructions, the agent naively assigns the entire total to \text{Start of Week Past Due} and then erroneously adds the demand on top. Furthermore, because the verification step occurs only after the table is generated, the agent either forgets to execute the checklist entirely or hallucinates a successful check on its own incorrect numbers. Consequently, the double-counting error is never caught.

\paragraph{The \textsc{FederatedSkill} Success Mechanism.}
Under the exact same conditions, the \textsc{FederatedSkill} agent succeeds because its inherited library encodes the invariant as a constructive formula rather than a post-hoc checklist. Specifically, the \texttt{workload-capacity-planning} skill explicitly instructs the agent to compute the starting value dynamically: $\text{Calc Start} = X - \text{Demand}[\text{first\_period}]$. 

This formulation mathematically prevents the double-counting error from occurring. The skill documentation explicitly highlights this edge case across multiple sections:

\begin{promptbox}[title={workload-capacity-planning/SKILL.md}]
\begin{verbatim}
# line 50, "Resolve Initial State":
Carefully separate "Start of Period Past Due" 
from "Scheduled Demand" when the prompt gives a 
combined initial condition (e.g., "Start+Demand=X").
Avoid double-counting demand in the first period.

# line 77, "Initial Condition Parsing":
If given `Start_of_Period_Past_Due + Demand = X`, 
compute `Calc_Start = X - Demand[first_period]` 
to prevent double-counting.

# line 107, "Edge Cases":
Initial-condition statements often combine 
"Start of Period Past Due" with "Scheduled Demand"; 
parse carefully to avoid double-counting.
\end{verbatim}
\end{promptbox}

\paragraph{Evolution of the Constructive Rule.}
Tracing the federation logs reveals that this robust rule was not discovered by the Kimi client, but was authored and iteratively refined by peer clients. The semantic evolution of this rule is explicitly captured in the patch payload submitted by client $u_0$ (Qwen3.6-Plus) in Round~2:

\begin{databox}[title={Skill Patch Payload: Round 2 Diff (Qwen3.6-Plus)}]
\begin{verbatim}
--- workload-capacity-planning/SKILL.md
+++ workload-capacity-planning/SKILL.md
-  - Formula: backlog = combined_value - week_4_demand
+  - Formula: calc_start = combined_value - 
+             demand[first_period]
 ...
- - **Initial Condition Parsing**: ... compute 
-   `Calc Start = X - Demand[Week 4]`
+ - **Initial Condition Parsing**: ... compute 
+   `Calc Start = X - Demand[first_period]`
\end{verbatim}
\end{databox}

This generalized patch is the culmination of a multi-round, cross-client refinement process:
\begin{itemize}[leftmargin=14pt,topsep=2pt,itemsep=1pt]
\item \textbf{Round 0:} Client $u_0$ encounters a similar pitfall and establishes the initial parsing rule, but hard-codes the variable for that specific trial (\texttt{Demand[Week 4]}).
\item \textbf{Round 1:} Client $u_2$ (Kimi~K2.5) encounters a different task variant and contributes a \texttt{horizontal-data-parsing.md} reference. The server evolution agent successfully merges this new logic while preserving $u_0$'s initial-condition rule.
\item \textbf{Round 2:} Client $u_0$ encounters another variant, recognizes the rigidity of its previous rule, and submits the patch shown above to abstract the hard-coded weekly constraints into period-agnostic logic.
\end{itemize}

The critical logic that enabled the Kimi~K2.5 agent to succeed in Round~6 was thus authored and refined by Qwen3.6-Plus across earlier rounds. In contrast, the isolated SE agent, lacking exposure to peer trajectories, accumulated a bloated 207-line skill containing brittle post-hoc checks and failed. This trajectory-level evidence directly explains the substantial performance gap observed in the PCP / Kimi~K2.5 / \emph{Heterogeneous Backbone \& Harness} evaluation cell, validating the efficacy of cross-client capability transfer.

\subsection{Library Evolution Under SE and \textsc{FederatedSkill}}\label{app:case-B}

\paragraph{Skill Convergence and Line-Level Tracking.}
Building on the trajectory analysis of the Kimi~K2.5 pair, we examine how their underlying libraries evolved across the PCP task sequence. Although both methods ultimately develop a single primary skill for this family, they assign it different names (\texttt{excel-capacity-planning} for SE and \texttt{workload-capacity-planning} for \textsc{FederatedSkill}). Therefore, tracking the pure number of skills is insufficient for a direct comparison. Instead, we analyze the line count of \texttt{SKILL.md} over successive rounds to observe the structural library dynamics.

\begin{table}[h]
\centering
\resizebox{\columnwidth}{!}{\small
\begin{tabular}{lcc}
\toprule
Round & SE (Kimi~K2.5) & \textsc{FedSk} (Kimi~K2.5) \\
\midrule
0 & - & - \\
1 & - & 50 \\
2 & 103 & 71 \\
3 & 124 & 75 \\
4 & 152 & 84 \\
5 & 163 & 112 \\
6 & \textbf{201} & \textbf{112} \\
7 & 207 & 117 \\
8 & 207 & 134 \\
\midrule
Reward $(\Sigma)$ & \textbf{4/9} & \textbf{8/9} \\
\bottomrule
\end{tabular}}
\caption{Lines of \texttt{SKILL.md} per round on the matched PCP cell (SE skill: \texttt{excel-capacity-planning}; \textsc{FederatedSkill} skill: \texttt{workload-capacity-planning}). SE accumulates lines rapidly following failures (e.g., Rounds 6 to 7), while \textsc{FederatedSkill} maintains a stable size through systematic deduplication.}
\label{tab:caseB-lines}
\end{table}

\paragraph{Analysis of Skill Growth.}
Comparing the SE skill snapshots between Round~6 (failed) and Round~7 (successful) reveals three sources of unnecessary bloat: duplicated edge-case warnings (pasting verifier output under multiple headings), hardcoded task-specific constants from prior trials, and redundant checklist items. In contrast, the \textsc{FederatedSkill} library remains stable during the same period. The server evolution agent integrates only one new peer reference, leaving the core constructive rules intact.

\paragraph{Directory-Level File Dynamics.}
Although both methods yield a single primary \texttt{SKILL.md} file, their supporting \texttt{references/} directories diverge significantly in content and authorship (Table~\ref{tab:caseB-files}).

\begin{table}[h]
\centering
\small
\setlength{\tabcolsep}{6pt}
\resizebox{0.6\columnwidth}{!}{%
\begin{tabular}{ccc}
\toprule
Round & SE files & \textsc{FedSk} files \\
\midrule
2 & 2 & 4 \\
4 & 4 & 4 \\
6 & 4 & 4 \\
8 & 4 & 4 \\
\bottomrule
\end{tabular}}
\caption{Total files in the SE \texttt{excel-capacity-planning/} and \textsc{FederatedSkill} \texttt{workload-capacity-planning/} reference directories.}
\label{tab:caseB-files}
\end{table}

By Round~8, the \textsc{FederatedSkill} directory contains diverse reference files (e.g., \texttt{environment-setup.md}, \texttt{horizontal-data-parsing.md}, and \texttt{validation-checklist.md}) contributed by three different clients across early rounds. Meanwhile, the SE directory relies exclusively on files authored by a single isolated client. 

Ultimately, while a naive count of skill names suggests identical library sizes, the internal structures fundamentally differ. The isolated SE client appends text after each failure without consolidating prior content, accumulating redundant warnings and brittle constants. Conversely, the \textsc{FederatedSkill} evolution agent systematically deduplicates incoming patches against the existing library, maintaining a concise file under 135 lines for the entire run. Given the higher aggregate reward for \textsc{FederatedSkill} (8/9 compared to 4/9 for SE), this compactness confirms that the server evolution agent effectively eliminates redundancy without compromising capability.

\begin{table}[h]
\centering
\small
\resizebox{\columnwidth}{!}{
\begin{tabular}{@{}ll@{}}
\toprule
\textbf{Standard key} & \textbf{Hard-coded domain mapping} \\
\midrule
\texttt{removed\_ids}    & \texttt{dropped\_categories} (retail), \texttt{closed\_departments}, \\
                         & \texttt{retired\_assets}, \texttt{retired\_schools} \\
\addlinespace
\texttt{changed\_records}& \texttt{adjusted\_categories}, \texttt{updated\_departments}, \\
                         & \texttt{modified\_assets}, \texttt{revised\_schools} \\
\addlinespace
\texttt{added\_ids}      & \texttt{new\_categories}, \texttt{opened\_departments}, \texttt{new\_schools} \\
\bottomrule
\end{tabular}
}
\caption{Hard-coded domain mappings in \texttt{domain\_mapping.md}. No mapping exists for hardware/datacenter or pharmacy/medication terminology.}
\label{tab:case_overfit}
\end{table}

\subsection{Analysis of Performance Regressions}\label{app:case-regressions}

Although \textsc{FederatedSkill} yields a net positive transfer across most settings, performance drops occasionally occur. These regressions typically arise when a client's library becomes overly rigid after integrating peer contributions that hard-code prior schemas. The agent becomes biased toward historical templates, forcing patterns that succeeded on previous tasks into new sub-domains where they no longer fit. We dissect a representative single-cell regression: the Cross-Format-Data-Reconciliation (CFDR) family under the \emph{Heterogeneous Backbone \& Harness} setting for Kimi~K2.5 ($-12.5$\,pp).

\paragraph{Task-Level Breakdown and Failure Mode.}
The CFDR family contains eight cross-format reconciliation tasks. Both methods succeed on six identical tasks (cloud portfolio, hospital capacity, retail category, university funding, shipping manifest, course catalog), but \textsc{FederatedSkill} fails on Round~4 (05-datacenter-hardware-registry-diff) and Round~5 (06-hospital-medication-reconciliation). The verifier rejects the Kimi client's Round~4 output because the agent emits its diff under generic field names (\texttt{removed\_ids}, \texttt{changed\_records}) that ignore the task's domain-specific keys (\texttt{decommissioned\_servers}, \texttt{updated\_servers}).

\paragraph{Library Overfitting.}
Inspecting the Kimi client's library at the start of Round~4 reveals the structural root cause. Through cumulative absorption across Rounds 0--3, the merger has populated Kimi's working directory with a peer-authored \texttt{dataset-diff/references/domain\_mapping.md} that hard-codes field-name substitutions for the four sub-domains seen so far (Table~\ref{tab:case_overfit}).

\paragraph{Personalized Merge Pattern.}
A key observation is that the merger \emph{personalized} the library: \texttt{domain\_mapping.md} was synced to the Kimi client (\texttt{u2}) and the GLM client (\texttt{u1}, its author), but \emph{not} to the Qwen client (\texttt{u0}). The merged-patch provenance for Round 3 shows:

\begin{promptbox}[title={Round 3 merged\_for\_u0 (Qwen) vs merged\_for\_u2 (Kimi)}]
\begin{verbatim}
merged_for_u0 (Qwen) — upsert_paths:
  dataset-diff/SKILL.md                          (top=u1)
  dataset-diff/references/pdf_excel_workflow.md  (top=u2)
  # no domain_mapping.md

merged_for_u2 (Kimi) — upsert_paths:
  dataset-diff/SKILL.md                          (top=u1)
  dataset-diff/references/pdf_excel_workflow.md  (top=u2)
  dataset-diff/references/domain_mapping.md      (top=u1)
\end{verbatim}
\end{promptbox}

When Round~4 introduces the datacenter-hardware schema, the Kimi agent recognizes a structurally similar task pattern (entity registry diff) and defaults to the most prominent reference in its library, the \texttt{domain\_mapping.md} table. Since this table contains no hardware/datacenter row, the agent silently falls back to the generic \texttt{removed\_ids}/\texttt{changed\_records} schema---the verifier expects \texttt{decommissioned\_servers}/\texttt{updated\_servers} and rejects the output. In contrast, the Qwen client---never given the domain-mapping reference---writes a fresh extractor that reads the exact field names from the task prompt and succeeds.

\paragraph{Mechanism Analysis.}
The server evolution agent accumulates schema- and tool-specific guidance over successive rounds. For task variants that fall within the union of absorbed schemas, this accumulation is uniformly beneficial. For a genuinely new sub-domain, however, a peer-contributed mapping table can act as an anchor that biases the agent away from inspecting the task prompt. The mechanism is the dual of the success pathway analyzed in \S\ref{app:case-M}: just as peer integration can equip a client with a constructive rule it would never have discovered alone, the same channel can also propagate rigid artifacts whose specific contents become liabilities outside their original sub-domain. 

A promising mitigation strategy is to enforce stricter library hygiene during the merge phase---for instance, requiring the merger to verify whether a hard-coded mapping table actually covers the incoming task's schema and, if not, suppress the reference rather than ship it. We leave this refinement to future work.

\section{Concrete Artifacts: The Evolution Agent and Two-Level Memory}\label{app:artifacts}

This section details the concrete artifacts of the server-side evolution agent: the system skills governing its execution and the two-level capability memory it maintains. The abstract pipeline is described in Section~\ref{sec:method-server}; the exact artifacts provided below are drawn from a federated run under the \emph{heterogeneous backbone \& harness} setting.

\subsection{Stage 1: Evolution Planning and Shared Memory}
The evolution pipeline begins with a family-shared planning pass. The evolution agent executes the \texttt{task-update} skill to construct the high-level, family-shared capability matrix (\texttt{task\_memory.md}), which formalizes the collective task frontier and tracks unsolved workflow gaps for the entire federation.

\begin{promptbox}[title={System Skill: task-update}]
\begin{verbatim}
---
name: task-update
description: Cloud-side step 1 - maintain
  task_memory.md as the family's running list of
  tasks NOT YET adequately covered by the libraries.
---

# task-update

You do not touch any library. Your only output is an
updated task_memory.md.

## What task_memory.md is
A per-worker coverage matrix (not one covered flag):
- Each entry describes a task observed in the family,
  plus which workers (models) have it covered and
  which still have a gap.
- Coverage is per-worker: a task that worked for u1
  (glm-5) is NOT automatically covered for u0
  (qwen3.6-plus). Libraries and models differ.
- An entry is removed only when EVERY worker covered.

## Reward semantics
- reward = 1.0  -> covered (task passed)
- reward < 1.0  -> NOT covered (failed); fractional
  value is only a partial-progress hint, never
  "half success".
- Reward is per-(worker, model, task): u1=1.0 does
  not cover u0 or u2.

## Workflow
1. Read task_memory.md, library_skills.md, every
   patches/<wid>/meta.json and body.
2. Update/add task buckets
   (INPUT / TRANSFORMATION / OUTPUT).
3. Update each worker's cell from this round.
4. Update per-worker findings (model-specific
   patterns, >=2 rounds of evidence).
5. Hard cap: 100 lines. Write DONE.txt.
\end{verbatim}
\end{promptbox}

The family-shared \texttt{task\_memory.md} generated by this skill (captured here from the HWPX-Document-Automation family at Round~3, the same round whose Stage~2 execution trace appears below) maps out the global capability boundary:

\begin{memorybox}[title={Shared Memory: task\_memory.md}]
\begin{verbatim}
# task_memory.md - HWPX-Document-Automation
Round 3. Hancom Office document automation.

## Task buckets (concrete, observed)

### B1: HWPX Template Placeholder Fill with Value
        Preprocessing + Multi-Section (R0-R3)
- INPUT:
  - .hwpx template file (ZIP archive of HWPML XML)
  - JSON mapping {{placeholder}} -> raw values
    (some require preprocessing)
- TRANSFORMATION:
  1. Extract .hwpx as ZIP (Python zipfile, not
     shell unzip)
  2. Enumerate and process ALL Contents/section*.xml
     files - R3 revealed multi-section documents
     are common
  3. Parse each section for {{key}} inside <hp:t>
  4. Preprocess values (R2): Korean age, phone
     normalization, date formatting
  5. Replace placeholders preserving surrounding XML
  6. CRITICAL: remove <hp:linesegarray>...</> OR
     self-closing <hp:linesegarray /> from modified
     <hp:p> paragraphs only
  7. Repackage as ZIP_DEFLATED, preserving ZipInfo
- OUTPUT: valid .hwpx, zero {{...}} across all
  sections, correct ZIP structure
- Common failures:
  - forgot linesegarray removal (incl. self-closing)
  - only processed section0.xml, ignored sectionN+
  - skipped preprocessing -> wrong age/phone values

## Coverage matrix

| Bucket | u0(qwen) | u1(glm-5)        | u2(kimi) |
|--------|----------|------------------|----------|
| B1     | covered  | gap (r=0.66)     | covered  |

All workers passed R2. R3 revealed multi-section
requirement: u0 and u2 updated skills successfully
(r=1.0). u1's patch had r=0.66 (task did NOT pass).

## Per-worker findings

- u0 (qwen3.6-plus): stable, do not over-engineer.
- u1 (glm-5):
  - R3 patch r=0.66 (task did NOT pass). Patcher
    summary: "agent had to write custom code to
    process all sections" - the documented workflow
    alone was insufficient to drive the agent.
  - Pattern: u1's SKILL.md is verbose (workflow,
    examples, troubleshooting prose) but its
    script-level implementation has gaps, so the
    agent at trial time falls off the documented
    rails.
- u2 (kimi-k2.5): stable, do not over-engineer.

## What's missing (for step-2 mergers)

- u1 needs concrete multi-section script support.
  Apply u0/u2's helper script changes (auto-discovery
  of section*.xml) to u1's library. Do NOT copy u1's
  verbose prose style into u0/u2 - their concise
  style fits their models.
- All three workers now agree on core rules: multi-
  section processing, self-closing <hp:linesegarray />,
  preprocessing before replacement. Merge script-level
  fixes; documentation divergence is acceptable.
\end{verbatim}
\end{memorybox}

\subsection{Stage 1.5: Per-Trial Patch Distillation}
Before any server-side merging takes place, the local patcher running on each client distills the agent's most recent trial into a structured update. This patch serves as the fundamental communication unit consumed by the evolution agent in Stage~2. The patcher executes via a single LLM call utilizing the client's native backbone, ensuring the patch inherently reflects the client's stylistic preferences. The output consists of a lightweight JSON manifest accompanied by the specific files proposed for library upsertion.

The complete patch produced by client $u_0$ (Qwen3.6-Plus / Qwen Coder) following the HWPX-Document-Automation Round~3 trial includes the following manifest:

\begin{databox}[title={Patch Manifest: WorkerPatch JSON}]
\begin{verbatim}
{
  "worker_id": "u0",
  "reward": 1.0,
  "summary": "Updated hwpx-template-fill to handle
    multi-section HWPX files (placeholders often span
    section0.xml and section1.xml) and fixed a critical
    regex bug where self-closing <hp:linesegarray />
    tags were not matched, which would cause layout
    corruption. Enhanced SKILL.md with explicit multi-
    section processing rules and verification steps.",
  "delete_paths": []
}
\end{verbatim}
\end{databox}

The accompanying upserted file (the complete \texttt{hwpx-template-fill/SKILL.md} body emitted by the patcher) demonstrates how trial-and-error lessons are translated into a structured, model-specific skill document:

\begin{databox}[title={Patch Payload: SKILL.md Upsert}]
\begin{verbatim}
---
name: hwpx-template-fill
description: Fill templates in .hwpx (Hancom Office)
  documents by replacing {{...}} placeholders with
  values from JSON. Use when given an .hwpx template
  and a JSON mapping of field names to values. Handles
  ZIP extraction, multi-section XML text replacement,
  and layout cache invalidation.
---

# Fill HWPX Template

HWPX files are ZIP archives containing XML content.
To fill placeholders safely:

## Workflow
1. Inspect structure: Use Python zipfile to list
   contents. Main text lives in Contents/section0.xml,
   Contents/section1.xml, etc. Always process all
   section*.xml files, as placeholders may span
   multiple sections.
2. Identify placeholders: Look for {{key}} patterns
   inside <hp:t> tags across all sections.
3. Preprocess Data (if needed): scripts/fill_hwpx.py
   performs direct 1:1 replacement. If values require
   computation (age, phone normalization, conditional
   text, metadata), transform the JSON or build a
   replacement dictionary before applying it to XML.
4. Replace & Invalidate Cache:
   - Replace {{key}} with values.
   - Critical: Remove <hp:linesegarray>...</hp:linesegarray>
     or <hp:linesegarray /> from any modified <hp:p>
     element. HWPX caches layout coordinates here;
     leaving them causes overlapping/garbled text when
     string lengths change.
5. Repackage: Write modified XML back into a new ZIP
   with ZIP_DEFLATED compression, preserving original
   ZipInfo metadata and file order.
6. Verify: Open output ZIP, read all section XMLs,
   confirm zero {{...}} patterns remain.

## Verification Notes
- linesegarray elements will remain on static/unmodified
  paragraphs. This is expected and correct.
- Only verify that paragraphs containing replaced
  values do not contain linesegarray.

## Known invariants (by sub-task)
### B1: HWPX Template Placeholder Fill
- <hp:linesegarray> must be removed from any modified
  <hp:p>. Failure causes overlapping/garbled chars.

## Anti-Patterns
- Do not treat .hwpx as a plain text file. It is a
  binary ZIP archive.
- Do not skip <hp:linesegarray> removal.
- Do not rely on external HWP libraries unless
  necessary; zipfile + re is sufficient.
- Do not assume placeholders only exist in
  section0.xml. Always scan all section*.xml.

## Automation
Run scripts/fill_hwpx.py <template.hwpx> <data.json>
<output.hwpx> for direct 1:1 placeholder replacement
across all sections.
\end{verbatim}
\end{databox}

The patch additionally includes a companion helper script \texttt{scripts/fill\_hwpx.py} (80 lines, omitted for brevity) implementing the auto-discovery and \texttt{<hp:linesegarray>} removal operations. As detailed in the Stage~2 execution trace below, the evolution agent absorbs this entire patch wholesale into $u_0$'s library because the target client authored the successful patch natively.

\subsection{Stage 2: Personalized Library Evolution and Private Memory}
Following the planning pass, the agent executes the \texttt{merge-skill-patch} skill independently for each target client. This core skill consumes the Stage 1 coverage matrix alongside a persistent, low-level private memory (\texttt{memory.md}) isolated per client. This separation decouples shared task logic from model-specific deployment and stylistic constraints.

\begin{promptbox}[title={System Skill: merge-skill-patch}]
\begin{verbatim}
---
name: merge-skill-patch
description: Cloud-side merge of peer workers'
  skill-library patches into a target worker's
  private library in personalized federated learning.
---

# merge-skill-patch

You are the cloud merger in personalized federated
skill-evolution. M workers each ran a task this
round and the patcher distilled each trial into a
WorkerPatch. Each worker has its own diverging
library. You decide the target worker's next library.

## Inputs (already in cwd)
  meta.json        target_worker, target_model, family
  task_memory.md   family coverage matrix (READ-ONLY)
  memory.md        your notes from prior rounds (R/W)
  patches/<wid>/   every worker's proposals (read-only)
  library/         target's library (R/W; the output)
  .baseline_library/  read-only snapshot for revert
  peer_libraries/<p>/ each peer's full library (read)
  scripts/         helper scripts (read-only)

## Principles (excerpt)
- Wholesale > synthesize. When the target's own patch
  (or a peer's) has reward = 1.0, apply it whole; do
  not rewrite SKILL.md by merging pieces.
- Extend before adding (umbrella-first). A new sub-
  domain defaults to extending an existing skill;
  adding a new skill needs justification.
- Hard cap: 4 skills per family (target 2-3). Going
  past it without a "genuinely different pipeline"
  justification is a merger bug.
- When in doubt, prefer the target's own patch (same
  model, same CLI) over a peer's reward=1.0 patch.

## Output - before exiting
1. library/ reflects your final decision.
2. bash scripts/validate_library.sh passes.
3. DECISIONS.md - one row per touched path
   (path | action | source | reward | reason).
4. memory.md updated for next round.
5. DONE.txt - one-line summary.
\end{verbatim}
\end{promptbox}

The associated private \texttt{memory.md} for client $u_0$ (Qwen3.6-Plus / Qwen Coder) accumulates backbone-specific formatting guidelines across rounds to guide personalization. Note how it captures the model's preference for concise procedures over the verbose styles favored by peers:

\begin{memorybox}[title={Private Memory: memory.md (Client $u_0$)}]
\begin{verbatim}
# memory - u0 (qwen3.6-plus) on HWPX-Document-Automation

## Model-specific insights (cumulative)
- Style fit: qwen3.6-plus does well with concise
  numbered-procedure SKILL.md (short steps, explicit
  anti-patterns, minimal prose). Verbose multi-
  section SKILL.md works for kimi/glm but confuses
  qwen.
- Script preference: short functional scripts
  (<=80 lines) with explicit argparse; regex fits
  qwen better than ElementTree.
- Verification clarity: an explicit "Verification
  Notes" section reduces false-alarm risk.
- Bug-fix absorption: when a peer fixes a bug (u1's
  linesegarray regex overlap), apply it directly;
  peer reward=1.0 is evidence.

## Library architecture
- 1 skill hwpx-template-fill (umbrella) + 2
  reference files + 2 scripts.
- Peer naming aligned: u0/u1/u2 all use
  hwpx-template-fill.
- R2 consolidated preprocessing; R3 added multi-
  section HWPX support.

## Key invariants learned
- Self-closing <hp:linesegarray /> handled with
  the full form.
- Multi-section HWPX: process ALL section*.xml.
- <hp:run> splitting: text spans runs; work at
  <hp:p>/<hp:t> level.

## Open questions
- Umbrella covers all observed variants
  (placeholder / direct / pattern).
\end{verbatim}
\end{memorybox}

\subsection{Stage 2 Execution Trace and Decisions}
Each Stage 2 evolution instance executes within an isolated sandbox initialized by a target configuration mapping (\texttt{meta.json}):

\begin{databox}[title={Execution Context: meta.json}]
\begin{verbatim}
{
  "target_worker": "u0",
  "target_model": "qwen3.6-plus",
  "target_cli": "qwen-code",
  "merger_model": "glm-5",
  "family": "HWPX-Document-Automation",
  "round": 3,
  "peers": ["u1", "u2"],
  "all_workers": {
    "u0": {"model": "qwen3.6-plus",
           "cli": "qwen-code"},
    "u1": {"model": "glm-5",
           "cli": "claude-code"},
    "u2": {"model": "openai/kimi-k2.5",
           "cli": "kimi-cli"}
  }
}
\end{verbatim}
\end{databox}

To prevent structural naming drift and align peer code components, the environment supplies \texttt{library\_skills.md} to summarize the target family's current library status:

\begin{memorybox}[title={Federation State: library\_skills.md}]
\begin{verbatim}
# library_skills - each worker's current library
## u0 (model: qwen3.6-plus)
- hwpx-template-fill: Fill .hwpx templates by
  replacing {{...}} placeholders with values from
  JSON. Handles ZIP extraction, XML text
  replacement, and layout cache invalidation.
## u1 (model: glm-5)
- hwpx-document-processing: Process HWPX files -
  extract, modify, create HWPX archives. For
  Korean document templates / Hancom placeholders.
## u2 (model: openai/kimi-k2.5)
- hwpx-template-fill: Fill HWPX templates by
  replacing {{placeholders}} from JSON or a dict.
  For Korean doc automation, mail-merge, forms.
\end{verbatim}
\end{memorybox}

The evolution pass concludes by writing an auditable decision log (\texttt{DECISIONS.md}), tracking exactly why peer patches were absorbed, adjusted, or rejected based on profile compatibility:

\begin{logbox}[title={Decision Log: DECISIONS.md}]
\begin{verbatim}
# DECISIONS.md - R3 Merge for u0

path: hwpx-template-fill/SKILL.md
  action: apply_target  source: u0  reward: 1.0
  vs_peers: match_peers (u2)
  reason: target's verified multi-section + self-
    closing linesegarray fix; u0's concise 37-line
    style fits qwen better than u2's 85-line one.
path: hwpx-template-fill/scripts/fill_hwpx.py
  action: apply_target  source: u0  reward: 1.0
  vs_peers: match_peers (u2)
  reason: regex-based; auto-discovers and processes
    all section*.xml files.
path: hwpx-document-processing/*
  action: reject  source: u1  reward: 0.66
  vs_peers: peer_only
  reason: u1's patch failed (0.66); verbose 92-line
    SKILL.md unsuited to qwen.

Summary: applied u0's patch wholesale (reward 1.0);
rejected u1 (0.66); did not absorb u2 (u0's own
patch already validated and fits qwen better).
\end{verbatim}
\end{logbox}

\section{Communication Cost Analysis}\label{app:communication_cost}

A common concern with federated training is the bandwidth required to exchange model state between clients. In parametric federated learning, each round transmits a full set of model parameters (or their gradients), which for modern foundation models ranges from hundreds of megabytes to tens of gigabytes. \textsc{FederatedSkill} instead exchanges only the skill patch (a lightweight JSON manifest and plain-text file updates), so the payload is several orders of magnitude smaller.

\paragraph{Measured Per-Family Patch Payloads.}
For every task family evaluated in the \emph{heterogeneous backbone} setting, we report the total on-disk patch bytes exchanged over the full federation process. This total includes all rounds, all 3 workers, and both upload and download payloads of the run that supplied the best performing cell. Each row in Table~\ref{tab:payload_same_cli} represents the actual measured patch payload.

\begin{table}[htb]
\centering
\small
\resizebox{\columnwidth}{!}{
\begin{tabular}{@{}lcc@{}}
\toprule
\textbf{Family} & \textbf{Rounds} & \textbf{Total Patch Payload} \\
\midrule
Cross-Format-Data-Recon.     & 8 & 125.5 KB \\
HWPX-Document-Automation     & 8 & 166.8 KB \\
Inventory-\&-Finance-Integ.   & 8 & 188.4 KB \\
Supply-Chain-Replenishment   & 9 & 131.9 KB \\
Compensation-Scenario-Mod.   & 8 & 280.7 KB \\
Distribution-Center-Auditing & 8 & 207.2 KB \\
DMAIC-Quality-Analysis       & 9 & 112.2 KB \\
Document-Fraud-Detection     & 8 & 213.6 KB \\
Embedded-Data-Repair         & 8 & 249.5 KB \\
Financial-Statement-Rolling  & 9 & 179.1 KB \\
Healthcare-Cost-Benefit-Ana. & 9 & 99.2 KB \\
Industry-Correlation-Analysis& 8 & 102.8 KB \\
Medical-Data-Standardization & 9 & 183.8 KB \\
OCR-Data-Extraction          & 8 & 224.1 KB \\
Operational-Recovery-Plan.   & 8 & 110.0 KB \\
PPT-Formatting-Optimization & 8 & 192.6 KB \\
Production-Capacity-Planning & 9 & 269.3 KB \\
Sales-Pivot-Analysis         & 8 & 183.2 KB \\
SEC-13F-Financial-Analysis   & 8 & 197.6 KB \\
Weighted-Risk-Assessment     & 8 & 160.8 KB \\
\midrule
\textbf{Total (20 families)} &   & \textbf{3.49 MB} \\
\bottomrule
\end{tabular}
}
\caption{Total measured patch payload per family in the \textbf{heterogeneous backbone} setting over the full federation run.}
\label{tab:payload_same_cli}
\end{table}

Per family, the federation exchanges between 99 KB and 281 KB of patches across the entire 8--9 round run. The whole benchmark transmits $\sim 3.5$ MB total.

\paragraph{Implications.}
This huge reduction in communication overhead yields key practical advantages over parametric federated learning. First, because the payload size is determined by the complexity of the procedural knowledge rather than the model's parameter count, the communication cost is completely decoupled from backbone scaling. Second, the sub-megabyte footprint makes skill federation highly edge-feasible, even over constrained networks. Finally, unlike the opaque gradient tensors transmitted in parametric FL, the plaintext patch payload is fully auditable, making it uniquely suited for secure or regulated environments.

\section{Statistical Significance of Main Results}
\label{app:statistics}
We report a robustness analysis for the headline result in Table~\ref{tab:mixed-cli-se-vs-skf} under the heterogeneous backbone setting, comparing the SE baseline with \textsc{FederatedSkill} on the 5 representative families used in section~\ref{sec:ablation}.

To account for baseline variance, we evaluate the SE setup across multiple independent runs: 4 for Qwen3.6-Plus, 3 for GLM-5, and 2 for Kimi-K2.5. We then compare these aggregated baseline averages directly against the \textsc{FederatedSkill} scores from Table~\ref{tab:mixed-cli-se-vs-skf}.

\begin{table}[htb]
\centering
\small
\resizebox{\columnwidth}{!}{
\begin{tabular}{@{}lccc@{}}
\toprule
\multicolumn{4}{@{}l}{\textbf{Panel A: Per-family paired results}} \\
\midrule
\textbf{Family} & \textbf{Qwen3.6-Plus} & \textbf{GLM-5} & \textbf{Kimi-K2.5} \\
\midrule
PCP
& 66.7 / 77.8 (+11.1)
& 33.3 / 66.7 (+33.3)
& 11.1 / 44.4 (+33.3) \\
OpRec
& 25.0 / 37.5 (+12.5)
& 25.0 / 37.5 (+12.5)
& 25.0 / 37.5 (+12.5) \\
S13F
& 37.5 / 37.5 (+0.0)
& 25.0 / 50.0 (+25.0)
& 12.5 / 25.0 (+12.5) \\
ICA
& 62.5 / 87.5 (+25.0)
& 100.0 / 100.0 (+0.0)
& 87.5 / 100.0 (+12.5) \\
OCR
& 50.0 / 50.0 (+0.0)
& 62.5 / 75.0 (+12.5)
& 62.5 / 75.0 (+12.5) \\
\midrule
\multicolumn{4}{@{}l}{\textbf{Panel B: Aggregate statistics}} \\
\midrule
\textbf{Backbone} & \textbf{$\Delta$} & \textbf{Std} & \textbf{95\% CI; W/T/L} \\
\midrule
Qwen3.6-Plus & +9.76 & 10.39 & [+2.22, +17.50]; 3/2/0 \\
GLM-5        & +17.07 & 12.84 & [+7.50, +25.83]; 4/1/0 \\
Kimi-K2.5    & +16.66 & 9.30  & [+12.50, +25.00]; 5/0/0 \\
\midrule
\textbf{Pooled (All)} & \textbf{+14.35} & \textbf{10.68} & \textbf{[+9.17, +19.62]; 12/3/0} \\
\bottomrule
\end{tabular}
}
\caption{Robustness analysis on the 5 representative families. Panel A reports SE baseline / \textsc{FederatedSkill} pass rates and the paired gain for each family. Panel B summarizes aggregate statistics across families.}
\label{tab:robustness_appendix}
\end{table}

As shown, \textsc{FederatedSkill} improves over the SE baseline by an average of $+14.35$\,pp across the 15 paired configurations, with a 95\% bootstrap confidence interval of $[+9.17, +19.62]$. The strictly non-negative W/T/L profile, with 12 wins, 3 ties, and 0 losses, provides additional evidence that collaborative skill evolution consistently improves agent capability without inducing regressions on this representative subset.

\section{Privacy Audit}
\label{sec:privacy}

A central design claim of FederatedSkill (\S\ref{sec:problem-formulation}) is that communicating only the LLM-distilled semantic skill patch $\delta_i^t = g_i(\mathcal{L}_i^t, \mathcal{B}_i^t, \rho_i)$, rather than the raw trajectory batch $\mathcal{B}_i^t$, preserves privacy by construction. We test this claim empirically via a comprehensive attribute-inference audit across multiple task families on the $N{=}3$ homogeneous GLM-5 run.

\subsection{Threat Model}
\label{sec:privacy-threat}

We model the server-side evolution agent $\mathcal{M}$, and any client receiving a personalized update $\Delta_i^t$, as honest-but-curious. The adversary's observation set is the round-$t$ outgoing message, $O_i^t = (\rho_i, \delta_i^t)$, together with every file the patch upserts into the local library $\mathcal{L}_i^t$. The adversary does \emph{not} observe the underlying trajectory $\tau \in \mathcal{B}_i^t$, the tool I/O, or the task instance $x \sim \mathcal{D}_i$. For a candidate sensitive entity $e$ extracted from $\tau$, the adversary's goal is to infer whether $e \in \tau$. We evaluate privacy using the \emph{Sensitive Entity Leakage Rate} (SELR):
\begin{equation}
\mathrm{SELR}(C) = \frac{\bigl|\{\, e \in \mathcal{E}_{\mathrm{sens}}(\tau) \,:\, \mathrm{leak}(e, C) \,\}\bigr|}{\bigl|\mathcal{E}_{\mathrm{sens}}(\tau)\bigr|},
\label{eq:selr}
\end{equation}
where $C$ represents the target corpus (e.g., the outgoing patch $\delta_i^t$). An entity is considered leaked if it exact-matches a substring of $C$ \emph{or} if independent LLM judges determine that $C$ reveals $e$. The core privacy claim is that $\mathrm{SELR} \to 0$ on the outgoing patch; any residual signal is quantified by the adversary's advantage relative to a randomly sampled, unrelated patch.

\subsection{Methodology}
\label{sec:privacy-method}

Our audit pipeline consists of three stages evaluated across sampled task families, clients, and federation rounds.

\paragraph{Stage 1: Entity Extraction.}
An LLM extractor parses the full trajectory $\tau$ and generates a set of candidate entities. Each entity is categorized by type (e.g., \texttt{file\_path}, \texttt{person\_name}) and assigned a sensitivity label: \emph{sensitive}, \emph{task\_necessary}, or \emph{neutral}. Domain-specific vocabulary necessary for skill generalization (e.g., ``SKU'', ``inventory'') is explicitly categorized as \emph{task\_necessary} and excluded from the leakage analysis. Headline SELR values are strictly reported on the sensitive subset.

\paragraph{Stage 2: Audit Corpora Assembly.}
For each sampled instance, we assemble multiple corpora for comparison: the \textbf{outgoing patch} ($\delta_i^t$ and upserted files); the \textbf{personalized update} ($\Delta_i^t$ returned by $\mathcal{M}$); the \textbf{task floor} (public baseline derived from the task name); a \textbf{strawman summary} (a naive, verbatim LLM summarization of $\tau$ without abstraction directives, serving as an upper bound for leakage); and a \textbf{cross-family patch} (a negative control drawn from an unrelated family).

\paragraph{Stage 3: Evaluation Rule.}
Leakage is determined using a strict-\textsc{and} protocol: an entity $e$ is flagged as leaked if it passes a numeric-safe exact-match test against the corpus, \emph{or} if two independent LLM judges from different provider families agree that the semantic context reveals the entity.

\subsection{Results}
\label{sec:privacy-results}

\begin{table}[t]
\centering
\small
\setlength{\tabcolsep}{6pt}
\resizebox{\columnwidth}{!}{%
\begin{tabular}{lr@{\,}l}
\toprule
Target corpus $C$ & \multicolumn{2}{c}{SELR$(C)$, strict, 95\% CI} \\
\midrule
raw trajectory (ceiling)        & 100.00 & \% \\
top-K strawman summary          &  52.00 & \% [49.1, 54.9] \\
public task description (floor) &  22.53 & \% [20.2, 25.1] \\
\textbf{outgoing skill patch}   & \textbf{5.08}  & \textbf{\% [3.9, 6.5]} \\
\textbf{merged broadcast patch} & \textbf{0.09}  & \textbf{\% [0.02, 0.5]} \\
cumulative library (round-last) &   3.12 & \% [2.3, 4.3] \\
cross-family neg.\ (sanity)     &   0.09 & \% [0.02, 0.5] \\
\bottomrule
\end{tabular}}
\caption{Sensitive Entity Leakage Rate across audit surfaces.
$n_{\mathrm{sens}}{=}1{,}123$ pooled across 54 units (6 families
$\times$ 3 workers $\times$ 3 rounds). Strict-\textsc{and} rule
(exact-match \emph{or} both LLM judges agree); 95\% Wilson CIs.
}
\label{tab:selr-headline}
\end{table}

Table~\ref{tab:selr-headline} summarizes the SELR across audit surfaces. The outgoing patch $\delta_i^t$ leaks only $5.08\%$ of sensitive trajectory entities. Once aggregated, the personalized update $\Delta_i^t$ broadcast back to clients drops this leakage to a negligible $0.09\%$. In stark contrast, a naive top-$K$ strawman summary leaks $52.0\%$. Notably, the cross-family negative control sits at $0.09\%$, which is statistically indistinguishable from the personalized update, confirming that our audit pipeline does not systematically over-report. The adversary advantage, defined as $\mathrm{SELR}(\delta_i^t) - \mathrm{SELR}(\text{cross-family})$, is only $0.050$.

The comparison against the strawman summary highlights the efficacy of the reflection procedure $g_i$. By swapping a generic LLM summary for FederatedSkill's abstraction directive, SELR is reduced by an order of magnitude ($52.0\% \to 5.1\%$). This confirms that the privacy gains stem directly from $g_i$'s explicit instructions to suppress task-specific values, identifiers, and one-off outputs, rather than from inherent LLM compression.

\begin{figure}[t]
  \centering
  \includegraphics[width=0.95\linewidth]{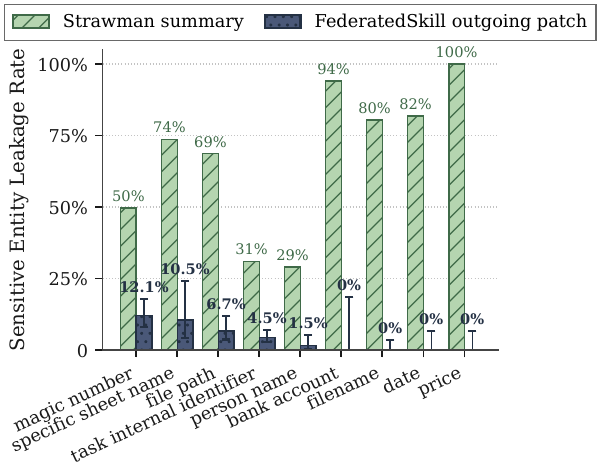}
  \caption{Sensitive Entity Leakage Rate into the outgoing patch $\delta_i^t$ (\textcolor[rgb]{0.29,0.34,0.47}{navy}) versus the top-K strawman summary (\textcolor[rgb]{0.25,0.42,0.28}{sage}), broken down by entity type. PII-grade categories (\texttt{filename}, \texttt{bank\_account}, \texttt{date}, \texttt{price}) exhibit $0\%$ leakage through the patch, whereas the strawman leaks $80$--$100\%$ on these same categories.}
  \label{fig:per-type}
\end{figure}

Furthermore, the per-type SELR breakdown (Fig.~\ref{fig:per-type}) reveals that residual leakage is highly concentrated in less critical categories. Severe PII-grade categories---such as \texttt{filename}, \texttt{bank\_account}, and \texttt{price}---leak at exactly $0\%$. The minor non-trivial leakage rates primarily originate from domain constants (e.g., \texttt{magic\_number} at $12.1\%$) or task-internal identifiers that occasionally surface in abstracted example snippets.

\subsection{Reconstruction Attack}
\label{sec:privacy-recon}

\begin{figure}[t]
  \centering
  \includegraphics[width=0.95\linewidth]{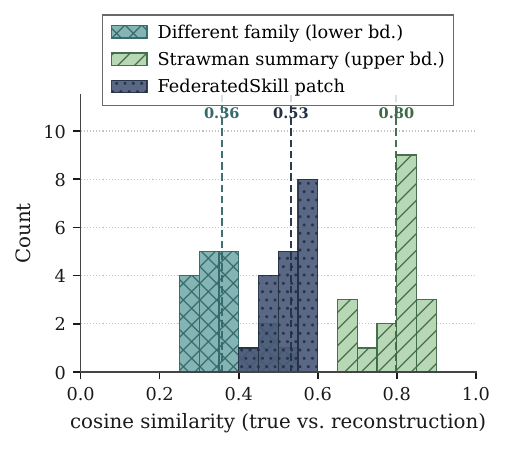}
  \caption{Task-reconstruction attack. Cosine similarity between the true task description and reconstructions derived from various sources. The patch-based reconstruction (navy) clusters near the lower bound (teal), demonstrating that semantic patches effectively obscure the original task shape.}
  \label{fig:recon}
\end{figure}

To verify that $\delta_i^t$ does not inadvertently leak the overarching \emph{shape} or intent of the task, we simulate a reconstruction attack. An LLM adversary is provided solely with $\delta_i^t$ (and its associated files) and prompted to reconstruct the original task description. The semantic similarity between the reconstruction and the true task yields a score of $0.532$, which aligns closely with the different-family lower bound ($0.357$) and falls significantly short of the strawman upper bound ($0.799$) (Fig.~\ref{fig:recon}). This demonstrates that reconstructing the task context from $\delta_i^t$ provides minimally more information than analyzing an unrelated patch.

\subsection{PII Canary Test}
\label{sec:privacy-canary}

Because the primary benchmark dataset contains no real Personally Identifiable Information (PII), we validate robustness via a direct canary injection test. Six synthetic PII canaries (e.g., formatted SSNs, internal employee IDs, confidential project codes) were injected into various locations across several trajectories (task prompts, reasoning traces, and tool observations). Each modified trajectory $\tau'$ was processed through the reflection procedure $g_i$ to generate a patch. 

\begin{table}[t]
\centering
\small
\setlength{\tabcolsep}{6pt}
\resizebox{\columnwidth}{!}{%
\begin{tabular}{llr}
\toprule
Canary value & Type & Leaked in \\
\midrule
\texttt{Marcia Doleba}                 & person & 0 / 9 \\
\texttt{Vinh Kostov-Reilly}            & person & 0 / 9 \\
\texttt{EMP-99317-XJ}                  & internal id & 0 / 9 \\
\texttt{412-77-8831}                   & SSN-shape & 0 / 9 \\
\texttt{mdoleba@acme-internal.test}    & internal email & 0 / 9 \\
\texttt{PROJ-CONFIDENTIAL-7Q}          & project code & 0 / 9 \\
\midrule
\textbf{Total}                         &                 & \textbf{0 / 54} \\
\bottomrule
\end{tabular}}
\caption{Stage 7 synthetic-PII canary survival. Each canary was injected
into the user task prompt, agent reasoning, and a tool observation of 9
Compensation-Scenario-Modeling trajectories; each modified trajectory
was re-distilled into a skill patch by an LLM patcher. ``Leaked in''
counts patches in which the canary survives either an exact substring
match or a Sonnet semantic-leak judge call. All 54 audits returned 0.}
\label{tab:canary}
\end{table}

Subsequent audits of these patches, utilizing both exact-substring matching and semantic judging, confirmed $0$ leaks across all canary trials (Table~\ref{tab:canary}). Every injected PII class was successfully and entirely suppressed by the abstraction process.

\section{Potential Risks and Broader Impact}
\label{sec:potential-risks}

While \textsc{FederatedSkill} is designed to enhance the capability and efficiency of productive LLM agents in a privacy-preserving manner, we acknowledge several potential risks and broader impacts associated with its real-world deployment:

\paragraph{Dual-Use and Malicious Skill Evolution.} As a general-purpose framework for collaborative agent self-improvement, the semantic patch-sharing protocol could theoretically be adapted for malicious objectives. For instance, adversarial actors might deploy a similar federated loop to collaboratively evolve networks of agents optimized for executing automated cyberattacks, bypassing web-security filters, or coordinating sophisticated phishing campaigns. Countering this risk necessitates strict access controls on the evolution server and the development of proactive patch-screening protocols.

\paragraph{Misplaced Trust in Empirical Privacy.} Although our rigorous privacy audit (\S\ref{sec:privacy}) demonstrates that semantic abstraction reduces sensitive entity leakage to near-baseline levels ($0.09\%$ in the personalized update), \textsc{FederatedSkill} relies on empirical validation rather than hard cryptographic guarantees (such as formal Differential Privacy). There is a minor risk that deployment organizations might overestimate the system's privacy boundaries, leading to a false sense of absolute security when handling highly regulated or extreme-sensitivity data without additional encryption layers.

\paragraph{Bias Amplification and Behavior Homogenization.} Aggregating skill patches across multiple independent clients risks propagating latent behavioral biases or flawed heuristics from dominant users to the rest of the federation. If specific client agents frequently upload patches reflecting biased reasoning paths, a server-side aggregation agent without explicit fairness or algorithmic alignment constraints might inadvertently standardize these anti-patterns, leading to capability homogenization and the spread of suboptimal workflows across benign clients.

\section{Artifact Licenses and Intended Use}
\label{sec:artifact-licenses}

In this work, we utilize the SkillFlow benchmark~\citep{zhang2026skillflow} alongside three large language models: Qwen3.6-Plus, GLM-5, and Kimi K2.5. We ensure that our usage of these artifacts strictly adheres to their respective licenses and terms of service. 

Specifically, the SkillFlow benchmark is utilized solely for academic evaluation purposes in accordance with its open-source research license. Access to the backbone models is governed by the API terms and acceptable use policies provided by Alibaba Cloud (Qwen), Zhipu AI (GLM), and Moonshot AI (Kimi). Our usage of these models to conduct automated agent task executions, extract semantic skill patches, and evaluate collaborative evolution is strictly for non-commercial academic research. This is fully consistent with their intended use cases. 

Furthermore, we do not distribute any proprietary model weights or internal data from the API providers. The code, framework infrastructure, and the federated skill patches generated during this study will be open-sourced under an MIT license to facilitate future research in agent collaboration, aligning with standard community practices.

\end{document}